\RequirePackage[svgnames]{xcolor}
\documentclass[11pt,letterpaper]{teamreport}

\microtypesetup{activate=false}
\usepackage[comma,authoryear,compress]{natbib}
\usepackage{subcaption,mathtools,array,multirow,makecell,longtable,float,placeins,needspace,tikz}
\usetikzlibrary{arrows.meta,positioning,calc,fit}
\tcbuselibrary{breakable}

\newcommand{\real}{\mathrm{real}}
\newcommand{\simu}{\mathrm{sim}}
\newcommand{\argmin}{\operatorname*{arg\,min}}

\newcommand{\EconGym}{EconGym}

\newcolumntype{L}[1]{>{\raggedright\arraybackslash}p{#1}}
\newcolumntype{C}[1]{>{\centering\arraybackslash}p{#1}}

\newcounter{ewmalgorithm}
\newcounter{ewmline}
\newenvironment{ewmalgorithmbox}[1]{%
 \refstepcounter{ewmalgorithm}%
 \begin{tcolorbox}[
  breakable,
  colback=white,
  colframe=black!55,
  colbacktitle=black!6,
  coltitle=black,
  boxrule=.6pt,
  arc=0pt,
  left=5pt,right=5pt,top=4pt,bottom=4pt,
  before skip=10pt,after skip=10pt,
  fonttitle=\bfseries,
  title={Algorithm \theewmalgorithm: #1}]
 \small
 \setlength{\parskip}{2pt}
}{%
 \end{tcolorbox}
}
\newenvironment{ewmalgorithmsteps}{%
 \setcounter{ewmline}{0}%
 \begin{list}{}{%
  \setlength{\labelwidth}{1.8em}%
  \setlength{\labelsep}{.5em}%
  \setlength{\leftmargin}{2.3em}%
  \setlength{\itemsep}{0pt}%
  \setlength{\parsep}{0pt}%
  \setlength{\topsep}{2pt}%
 }%
}{%
 \end{list}
}
\newcommand{\algstep}{\refstepcounter{ewmline}\item[\theewmline:]}

\title{Economic World Model}
\runningtitle{Economic World Model}
\author[1,2]{Qirui Mi}
\author[3]{Yujia Ming}
\author[4]{Bo Li}
\author[5]{Haifeng Zhang}
\affil[1]{School of Computer Science and Artificial Intelligence, Shanghai University of Finance and Economics}
\affil[2]{Key Laboratory of Interdisciplinary Research of Computation \mbox{and Economics, Shanghai University of Finance and Economics}}
\affil[3]{School of Finance, Shanghai University of Finance and Economics}
\affil[4]{School of Economics, Peking University}
\makeatletter
\renewcommand*{\AB@affilsep}{\protect\qquad}
\makeatother
\affil[5]{Institute of Automation, Chinese Academy of Sciences}
\keywords{economic world models, multi-agent system, economic simulation, social simulation, counterfactual experiments}

\begin{document}
\begin{abstract}
Can artificial intelligence reproduce the economic dynamics that emerge as households, firms and governments interact and adapt through markets and institutions? To answer this question, we define an economic world model (EWM) as an agent's internal representation of how an economic system works, enabling counterfactual simulation and decision-making under resource constraints. We then propose a general framework for studying EWMs that links three tasks: simulating agent behaviour, learning economic transitions and systematically evaluating the resulting dynamics. As an initial implementation, we learn household policies from microeconomic data and calibrate a structural economic transition model using macroeconomic data, thereby linking individual decisions to aggregate dynamics. Our micro-to-macro evaluation framework compares simulated individual actions and group distributions with observed microdata and assesses macroeconomic dynamics through forecasting and counterfactual experiments. Experimental results show that simulated and observed household consumption distributions overlap by 94.9\%, while EWM forecasts closely track real GDP and several other aggregate levels. In government-borrowing counterfactuals, the EWM reproduces the benchmark life-cycle model's short-run transmission from borrowing to consumption. For ex ante planning, the EWM enables an LLM agent to achieve 10.29\% higher simulated GDP under the epidemic shock than direct LLM decisions. This initial EWM is only a starting point. We invite economists and AI researchers to build a stronger EWM together. Economists can contribute richer economic models and knowledge, while AI researchers can develop stronger data-driven methods for simulating agent behaviour and learning economic transitions. Together, we aim not only to reproduce real economic dynamics, but also to discover and test new mechanisms that support better decisions and benefit society.
\end{abstract}
\maketitle

\section{Introduction}

Rapid advances in artificial intelligence (AI) have reshaped economic research. Can this progress also help us better understand how economies work? To see where AI can contribute, we look back at how economists have modelled economic dynamics over the past several decades. Early macroeconometric models estimated how economic variables move together \citep{klein1955econometric,sims1980}. The Lucas critique pointed out that agents' responses to policy can alter these relationships, motivating dynamic stochastic general equilibrium (DSGE) models grounded in individual decisions and market equilibrium \citep{lucas1976econometric,kydland1982time,smets2007shocks}. Heterogeneous-agent New Keynesian (HANK) models bring differences across households into this framework \citep{kaplan2018monetary}. Agent-based computational economics (ACE) offers a complementary bottom-up approach, \mbox{simulating} how aggregate outcomes emerge from interactions among agents \citep{tesfatsion2006handbook,dawid2018agent}. AI brings new data-driven tools for modelling agent behaviour. Reinforcement learning trains agents to solve complex tasks through experience, while large language models (LLMs) can generate actions from context \citep{Zheng2022AIEconomist,Li2024EconAgent}. As heterogeneous agents interact through markets and institutions, their decisions change the economic conditions that shape their subsequent choices. The challenge is to use AI to model this evolving system well enough to reproduce the dynamics observed in real economies.

\begin{figure}[!t]
\centering
\includegraphics[width=.93\textwidth]{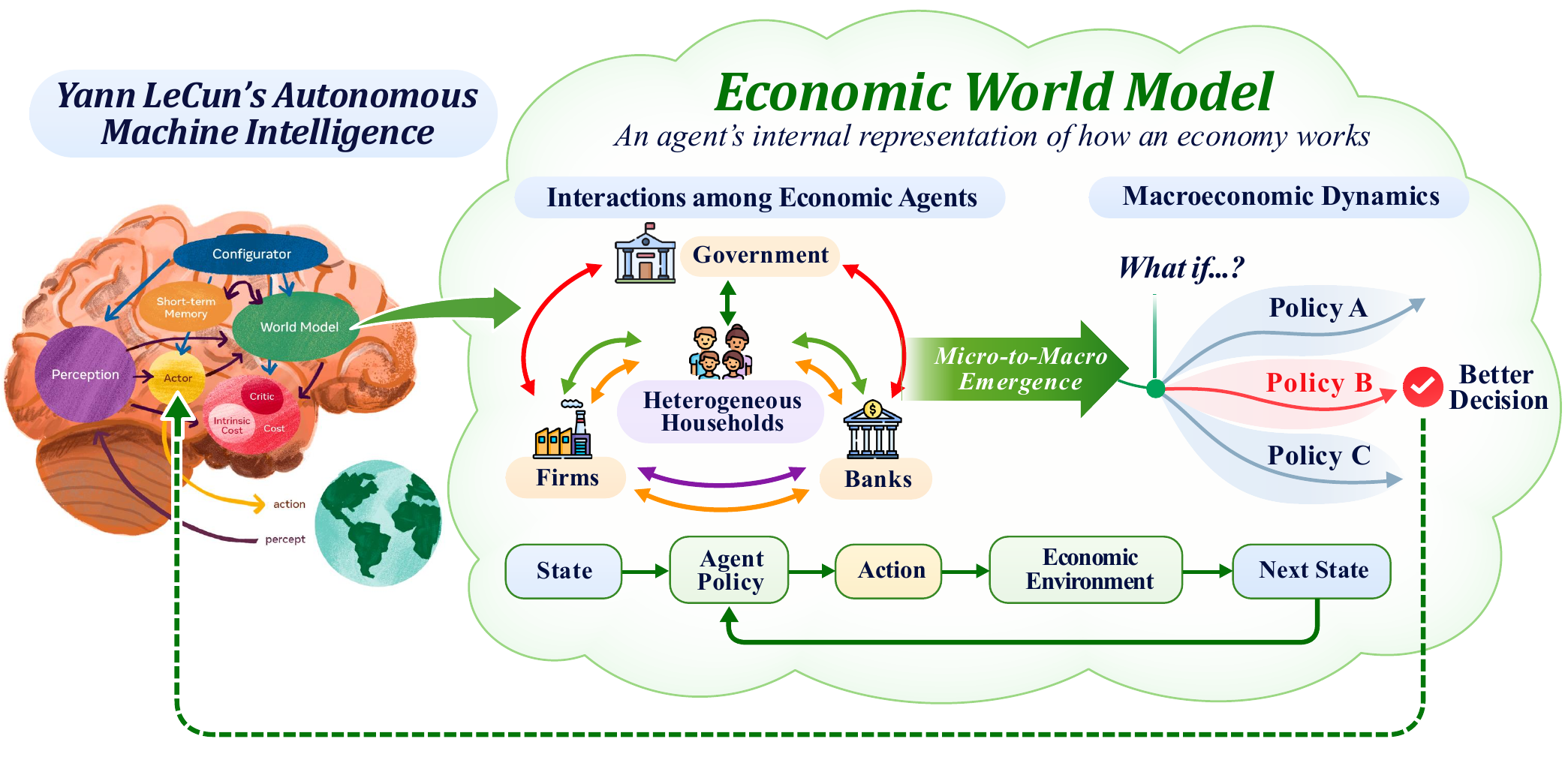}
\caption{\textbf{The economic world model framework.} Inspired by LeCun's autonomous intelligence architecture \citep{lecun2022path}, an EWM is an agent's internal model of how the economy works. It supports counterfactual experimentation and intelligent decision-making.}
\label{fig:ewm-overview}
\end{figure}

\Needspace{3\baselineskip}
To address this challenge, we define an \textbf{economic world model (EWM)} as an agent’s internal representation of how an economy works, enabling counterfactual simulation and decision-making under resource constraints (Figure~\ref{fig:ewm-overview}). To study EWMs, we propose a general multi-agent framework (Figure~\ref{fig:framework}, Section~\ref{sec:formulation}) that aims to reproduce real economic dynamics by minimising the discrepancy between simulated and real trajectory distributions. The framework links three tasks—simulating economic agent behaviour, learning economic transitions and systematically evaluating the resulting dynamics. At each step, agent policies map local observations to actions. The economic transition model maps the current state and the agents’ joint action to the next state, capturing how decisions propagate through markets and institutions. Repeating this cycle generates state–action trajectories that represent the economic dynamics we aim to predict. Our micro-to-macro evaluation framework (Section~\ref{sec:evaluation}) compares simulated individual actions and group distributions with microdata. At the macro level, forecasts assess how closely the EWM reproduces observed paths, while counterfactual experiments test whether shocks and policy changes propagate through economically plausible mechanisms.

The framework gives economists and AI researchers a common basis for extending an EWM. Economists can use structured economic models to specify how the economy evolves and extend the transition model to new settings. AI researchers can design new data-driven algorithms to learn economic transitions or better simulate how agents act and respond to changing economic conditions. Our micro-to-macro evaluation shows where the resulting dynamics agree with observations and economic benchmarks and where they diverge, guiding further refinements to agent policies and the transition model. We envision a shared EWM that draws on contributions from both communities to reproduce a wider range of real-world dynamics and support ex ante simulation and intelligent decision-making.

\Needspace{3\baselineskip}
As a starting point, we provide an initial implementation of our EWM framework (Section~\ref{sec:implementation}). Learning economic transitions from scratch with a purely data-driven generative model remains difficult, so we use structured models grounded in economic theory. To simulate household behaviour, we use behavioural cloning on microdata to capture individual action changes and group distributions of action levels and changes, laying the groundwork for counterfactual simulations. We use the trained agent policies and the transition model to advance the economy from one state to the next and generate simulated trajectories. We calibrate the transition model’s parameters by comparing these trajectories with observed macroeconomic paths. Because backpropagating the calibration loss through these multi-period rollouts is infeasible in this implementation, we develop Sobol–RBF, a derivative-free method for calibrating the transition model’s parameters. This initial implementation of the EWM yielded five main experimental findings (Section~\ref{sec:experiments}).

\Needspace{3\baselineskip}
\begin{enumerate}[leftmargin=*,label=(\arabic*),topsep=2pt,partopsep=0pt,parsep=0pt,itemsep=.2em]
 \item Simulated and observed distributions of household consumption levels overlapped by \textbf{94.9\%}. Individual adjustments, especially changes in hours worked, were harder to predict.
 \item The EWM’s one-year forecasts for 2024 were within \textbf{0.8\%} of observed output, consumption and productive capital levels. Abrupt turning points, such as the one triggered by the 2020 pandemic shock, were harder to predict.
 \item Given a pandemic shock to hours worked in 2020, the EWM reproduced the observed pattern of falling consumption and output and rising debt-to-GDP. When the shock ended in 2021, hours, consumption and output rebounded, and debt-to-GDP fell.
 \item The EWM reproduced some of the short-run responses to front-loaded borrowing in our implementation of \citeauthor{gertler1999government}’s~(\citeyear{gertler1999government}) life-cycle model, including higher consumption and lower total investment.
 \item By using the EWM to evaluate candidate tax policies ex ante, LLM agents achieved \textbf{10.29\%} higher simulated GDP in 2025 under the shock and \textbf{8.89\%} higher without it, relative to direct LLM decisions.
\end{enumerate}

\Needspace{12\baselineskip}
\noindent\textbf{Contributions.}\par
\begin{enumerate}[leftmargin=*,label=(\arabic*),topsep=2pt,partopsep=0pt,parsep=0pt]
 \item \textbf{Definition and general framework.} We define the economic world model and propose a general framework that links three tasks—simulating agent behaviour, learning economic transitions and systematically evaluating the resulting dynamics. The framework is extensible, allowing researchers to build a stronger EWM together by refining agent policies or transition models.
 \item \textbf{Implementation.} We combine a structured economic transition model with household policies learned from microdata through behavioural cloning and calibrate the transition parameters against macroeconomic data using a derivative-free Sobol–RBF search.
 \item \textbf{Evaluation and experimental findings.} We introduce a micro-to-macro framework for evaluating economic world models. Our evaluation shows that the initial EWM captures some macroeconomic trends and transmission patterns and helps LLM agents make better policy decisions in simulation. This implementation is a first step towards building more capable EWMs, with considerable room for improvement.
\end{enumerate}

\section{Related work}

This section reviews three strands of research on economic dynamics. It begins with econometric and equilibrium models, then turns to agent-based computational economics and AI-driven decision-making. Finally, it introduces economic world models as an emerging direction for understanding and predicting how economic systems evolve.

\subsection{Econometrics and equilibrium models}

Econometrics connects economic theory and observed data through probabilistic models and statistical inference \citep{haavelmo1944probability}. Early large-scale macroeconometric models represented economic dynamics through national-accounting identities and estimated behavioural equations, supporting forecasting and policy analysis \citep{klein1955econometric}. Vector autoregressions relaxed many a priori restrictions and characterised joint macroeconomic dynamics and responses to innovations \citep{sims1980}, while dynamic factor models summarised information from large time-series panels for forecasting \citep{stockwatson2002}. These approaches established empirical regularities and forecasting benchmarks, but policy counterfactuals require estimated relationships to remain stable across policy regimes. \citet{lucas1976econometric} argued that policy changes alter expectations and optimal decisions, so behavioural equations estimated under one regime may not remain valid under another. This critique strengthened the case for structural macroeconomic models that derive decision rules from preferences, technology, constraints and expectations \citep{lucas1976econometric,kydland1982time}.

In response, modern equilibrium-based macroeconomic models derive aggregate dynamics from agents' intertemporal decisions, resource constraints and market-clearing conditions \citep{lucas1976econometric,kydland1982time}. Real business cycle models provided an early template for dynamic stochastic general equilibrium (DSGE) models, combining optimising agents, stochastic productivity shocks and competitive equilibrium \citep{kydland1982time}. New Keynesian DSGE models incorporate nominal rigidities and real frictions to analyse business cycles and monetary transmission \citep{smets2007shocks}. Financial-accelerator models examine how credit frictions amplify these fluctuations \citep{bernanke1999financial}. Heterogeneous-agent models moved beyond representative-agent formulations by incorporating uninsurable idiosyncratic risk and wealth distributions into aggregate dynamics \citep{huggett1993risk,aiyagari1994uninsured,krusell1998income}. Heterogeneous-agent New Keynesian (HANK) models combined this heterogeneity with nominal rigidities to study how wealth, liquidity and marginal propensities to consume shape policy transmission \citep{kaplan2018monetary}. Solving heterogeneous-agent equilibrium models typically requires jointly determining individual policy functions, the evolving distribution of agents and equilibrium prices. Adding dimensions of heterogeneity, assets or shocks can make these models harder to solve, motivating computational approaches based on machine learning \citep{Hill2021SolvingHeterogeneousGE,Xu2023RegularizedCompetitiveEquilibria}.

\subsection{Agent-based computational economics and AI-driven decision-making}

Agent-based computational economics (ACE) offers another route for studying economic dynamics. In its classic formulation, ACE models economies as evolving systems of autonomous, interacting agents and studies how aggregate outcomes emerge from the bottom up \citep{tesfatsion2002bottomup,tesfatsion2006handbook}. Rather than solving an economy-wide equilibrium at every step, it specifies agent states, behavioural rules, interaction structures and institutions, then simulates their evolution over time \citep{epstein1999agent,tesfatsion2006handbook}. Early work showed how simple local rules can produce collective patterns \citep{schelling1971dynamic}; later models examined growth and business cycles \citep{dosi2010schumpeter} or explicitly accounted for flows and balance-sheet links among households, firms and banks \citep{caiani2016agent,dawid2018agent}. ACE thus makes the path from individual behaviour to aggregate dynamics explicit. How agents make decisions within that system remains a central modelling choice.

Reinforcement-learning agents learn policies from experience \citep{Sutton2018}. LLM agents can generate decisions from profiles, memory and language context \citep{Park2023GenerativeAgents}. This shift extends ACE's bottom-up framework beyond expert-specified rules by introducing learned policies and generative agents. Reinforcement learning has been applied to intertemporal household and macroeconomic decisions \citep{Shi2022LearningFromZero,Atashbar2023RLRBC}, pricing and market interaction \citep{calvano2020artificial}, tax policy with co-adaptive households and governments \citep{Zheng2022AIEconomist,Trott2021BuildingFoundation,Mi2023TaxAI}, and human-centred mechanism design \citep{Koster2022DemocraticAI}. A related approach formulates macroeconomic policy optimisation as a Stackelberg mean-field game \citep{Mi2024StackelbergMFG}. Simulation platforms such as \EconGym{} provide modular environments spanning broader economic roles and tasks \citep{mi2025econgym}. LLM agents have been used to simulate household and macroeconomic behaviour \citep{Li2024EconAgent,Douglas2024ConsumptionSavings}, financial markets and tax design \citep{Gao2024SimulatingFinancialMarket,Karten2025LLMEconomist}, and repeated firm competition \citep{Han2024GuineaPig,Fish2024AlgorithmicCollusion}. These studies focus on agent decision-making in simulated economic environments. Accurately predicting economic dynamics, however, also requires modelling how joint actions propagate through markets and institutions to shape future economic states.

\subsection{World models for economic dynamics}

In AI, a world model is an internal predictive representation of how an environment evolves under actions, supporting planning and policy learning through simulated futures \citep{ha2018world,lecun2022path}. Recent methods learn compact representations and transition dynamics from observations, then improve agent policies through imagined trajectories \citep{hafner2025mastering}. Economics needs this capability to represent individual behaviour and the aggregate dynamics arising from heterogeneous agents interacting through markets and institutions.

Economic world models remain an emerging research direction. \citet{cong2025economic} frames them as models of agents, environments and interactions specified by theory, learned from data or both. Building on this work, \citet[Sec.~1]{han2026agenticeconomies} describe an EWM system as ``a generative engine for economic environments''. We distinguish these roles from the agent's perspective. An economic environment provides feedback on an agent's actions, whereas an economic world model serves as the agent's internal model for simulating the consequences of potential actions before acting.

\begin{samepage}
To clarify the concept of an economic world model, we define:
\begin{quote}
An \textbf{economic world model} is an agent's internal representation of how an economic system works, enabling it to simulate potential economic outcomes under resource constraints, thereby supporting counterfactual experimentation and intelligent decision-making.
\end{quote}
\end{samepage}

However, several challenges make it difficult to apply existing world-model methods directly to economics. \textbf{Evolving economic mechanisms.} Changes in policy can alter expectations and behaviour, and institutions can evolve with the agents who interact within them \citep{lucas1976econometric,han2026agenticeconomies}. Empirical studies also document instability in macroeconomic time-series relationships \citep{stockwatson1996structural}. \textbf{Long time horizons.} Economic dynamics often unfold over long horizons, as the effects of underlying mechanisms take time to spread through the system. Monetary policy, for example, can affect the economy with lags of months or years \citep{burr2024transmission}. \textbf{Limited training data.} Video supplies dense sequences of consecutive observations, allowing world models to learn frame-to-frame physical transitions \citep{ha2018world}. Economic dynamics, by contrast, emerge from interactions among households, firms, markets and institutions, but economic datasets rarely capture these interactions jointly. Household data, for example, may record consumer behaviour without matching observations of how firms and markets respond to the same event. As a result, coherent trajectories linking agents, markets and time are scarce, making it difficult to learn economic dynamics at scale.

Given these challenges, learning economic dynamics from scratch with a generative model remains difficult. We therefore combine established economic knowledge with data-driven AI as a practical route to an initial economic world model. Our general framework represents the economy as a multi-agent system and uses economic knowledge to structure its transition model. Data-driven methods learn agent behaviour and calibrate transition parameters against real data. The next section presents this framework.

\section{A general framework for economic world models}
\label{sec:formulation}

\subsection{A partially observable multi-agent economic system}

Consider an economy in discrete time, $t\in\{0,\ldots,T\}$. The set of active economic agents at time $t$ is $\mathcal I_t=\{1,\ldots,N_t\}$, where $N_t$ denotes the number of active agents. We model governments, households, firms and banks as agents that make their own decisions within a common economic system.

Let $\mathbf s_t\in\mathcal S$ denote the global state vector representing the economy at time $t$. This vector can include macroeconomic variables (e.g., the price level, real GDP and inflation) and microeconomic states (e.g., household assets and wealth, and firm and institutional balance sheets). We model the economic system as a partially observable Markov game (POMG) \citep[Sec.~2]{liu2022samplepomg}. Under the Markov assumption, the next-state distribution is independent of the preceding state--action history given the current state $\mathbf s_t$ and joint action $\mathbf a_t$.
\begin{equation}
 P^{\real}(\mathbf s_{t+1}\mid \mathbf s_{0:t},\mathbf a_{0:t})
 =P^{\real}(\mathbf s_{t+1}\mid \mathbf s_t,\mathbf a_t).
 \label{eq:markov}
\end{equation}

Each agent receives only a private, local observation of $\mathbf s_t$. This observation includes agent-specific attributes, such as type, preferences and initial endowments, together with the current economic information available to that agent. For agent $i$, let $O_i$ denote its observation kernel and $O$ the joint observation kernel, with $O_i$ as its marginal. Let $\mathbf o_t=(o_t^i)_{i\in\mathcal I_t}$ denote the joint observation vector. The individual and joint observations follow
\begin{equation}
 o_t^i\sim O_i(\cdot\mid \mathbf s_t),\qquad
 \mathbf o_t\sim O(\cdot\mid\mathbf s_t),
 \label{eq:observation}
\end{equation}

Given its local observation $o_t^i$, agent $i$ selects an action according to its decision policy $\pi^i$. The aggregation rule $G$ then collects the individual actions into the joint action vector $\mathbf a_t$. Together, the individual policies and $G$ induce the joint policy $\Pi$, which gives the conditional distribution of $\mathbf a_t$ given $\mathbf o_t$:
\begin{equation}
 a_t^i\sim\pi^i(\cdot\mid o_t^i),\qquad
 \mathbf a_t=G\!\left(a_t^1,\ldots,a_t^{N_t}\right),\qquad
 \mathbf a_t\sim\Pi(\cdot\mid\mathbf o_t),
 \label{eq:aggregation}
\end{equation}

Once all agents have made their current decisions, their joint action $\mathbf a_t$ drives the transition to the next economic state:
\begin{equation}
 \mathbf s_{t+1}\sim P(\cdot\mid \mathbf s_t,\mathbf a_t).
 \label{eq:environment-transition}
\end{equation}

After $\mathbf s_{t+1}$ is generated, the cycle of observation, action and transition repeats from equation~\eqref{eq:observation} until a termination condition is met. This recursion generates the state--action trajectory $\tau=(\mathbf s_0,\mathbf a_0,\mathbf s_1,\ldots,\mathbf a_{T-1},\mathbf s_T)$. Joint observations $\mathbf o_t$ are intermediate variables used to generate actions and are therefore not included in $\tau$. Each step of the trajectory can be summarised as
\begin{equation}
 \mathbf s_t\longrightarrow \mathbf o_t
 \xrightarrow{\Pi}\mathbf a_t
 \xrightarrow{P}\mathbf s_{t+1}
 \longrightarrow\cdots\longrightarrow \mathbf s_T.
 \label{eq:one-step-chain}
\end{equation}

\begin{figure}[!t]
\centering
\includegraphics[width=\textwidth]{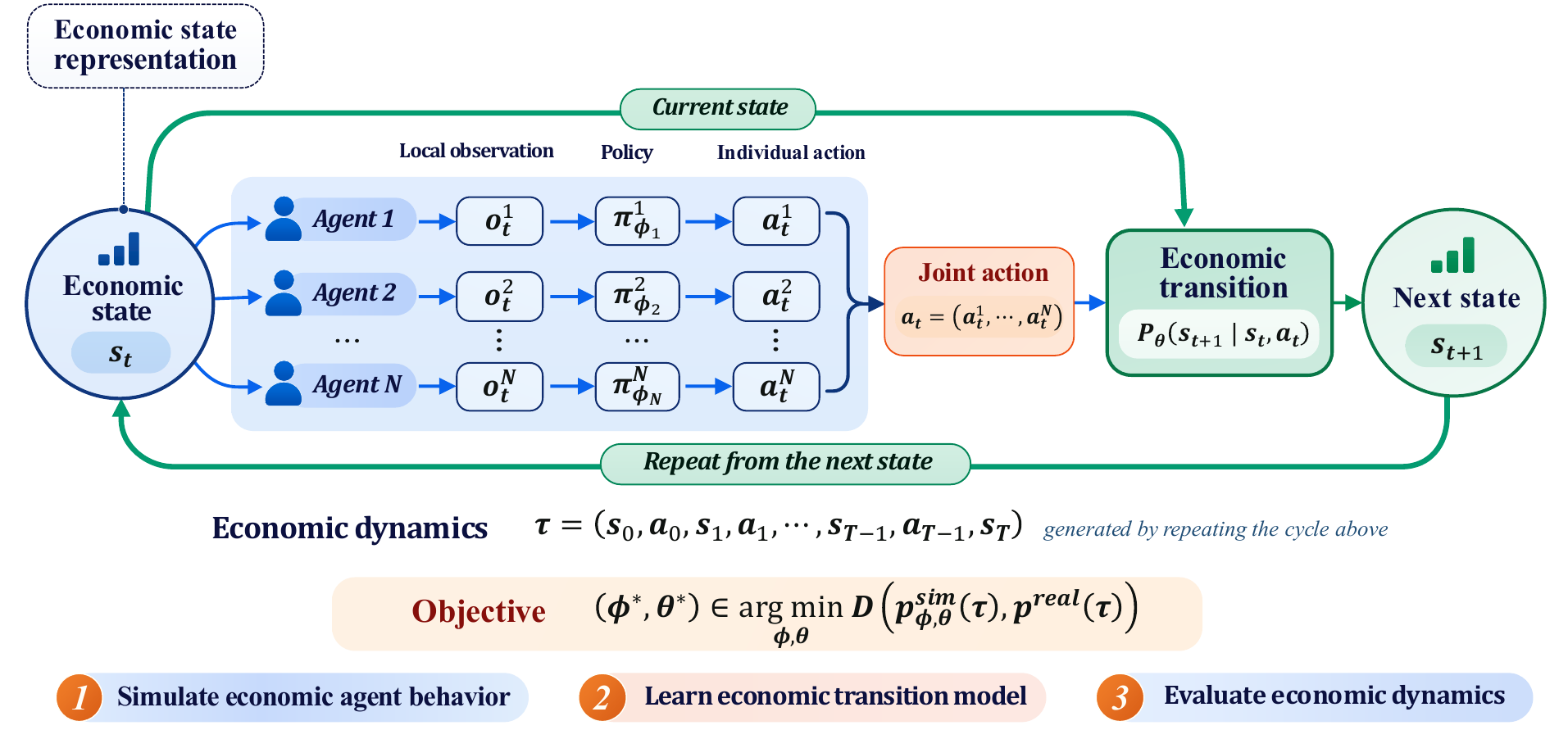}
\caption{\textbf{General framework of the Economic World Model.} At each step, agents act on their own local observations of the economy. The economic transition model uses the current state and the agents' joint action to generate the next state. Repeating this cycle yields state--action trajectories that describe the economy's dynamics. The objective is to minimise the discrepancy between simulated and real economic dynamics.}
\label{fig:framework}
\end{figure}

\Needspace{8\baselineskip}
\subsection{Optimisation objective of the economic world model}

The economic world model aims to reproduce real economic dynamics as closely as possible. We parameterise agent $i$'s policy as $\pi_{\varphi_i}^i$ and collect the policy parameters in $\varphi=\{\varphi_i\}_{i\in\mathcal I_t}$. The joint policy $\Pi_\varphi$ is parameterised by $\varphi$, and the economic transition model $P_\theta$ is parameterised by $\theta$. Together they parameterise the same trajectory-generating process:
\begin{equation}
 \mathbf s_t\longrightarrow\mathbf o_t
 \xrightarrow{\Pi_\varphi}\mathbf a_t
 \xrightarrow{P_\theta}\mathbf s_{t+1}
 \longrightarrow\cdots\longrightarrow \mathbf s_T.
 \label{eq:parametric-economic-dynamics}
\end{equation}
For compact notation, we combine the joint observation kernel $O$ and the joint policy $\Pi_\varphi$ into a single state-conditional action distribution $\widetilde\Pi_\varphi$.
\begin{equation*}
 \mathbf o_t\sim O(\cdot\mid\mathbf s_t),\qquad
 \mathbf a_t\sim\Pi_\varphi(\cdot\mid\mathbf o_t)
 \quad\Longrightarrow\quad
 \mathbf a_t\sim\widetilde\Pi_\varphi(\cdot\mid\mathbf s_t).
\end{equation*}
For an initial-state distribution $p_0$, this recursion induces the following probability for any trajectory $\tau$:
\begin{equation}
 p_{\varphi,\theta}^{\simu}(\tau)
 =p_0(\mathbf s_0)\prod_{t=0}^{T-1}
 \widetilde\Pi_\varphi(\mathbf a_t\mid \mathbf s_t)
 P_\theta(\mathbf s_{t+1}\mid \mathbf s_t,\mathbf a_t).
 \label{eq:trajectory-density}
\end{equation}
Equation~\eqref{eq:trajectory-density} maps the policy and transition parameters $(\varphi,\theta)$ to the simulated trajectory distribution $p_{\varphi,\theta}^{\simu}(\tau)$. Let $p^{\real}(\tau)$ denote the corresponding distribution of real economic trajectories, and let $\mathcal D$ measure the discrepancy between the two distributions. The objective shown in Figure~\ref{fig:framework} is
\begin{equation}
 (\varphi^*,\theta^*)
 \in\argmin_{\varphi,\theta}
 \mathcal D\!\left(
 p_{\varphi,\theta}^{\simu}(\tau),
 p^{\real}(\tau)
 \right).
 \label{eq:ewm-objective}
\end{equation}

Equation~\eqref{eq:ewm-objective} defines the overall objective of matching simulated and real economic trajectory distributions. This objective raises three connected research questions:
\begin{enumerate}
 \item \textbf{Learning to simulate economic agent behaviour ($\varphi^*$).} How can we learn individual policies so that simulated decisions, strategic interactions and distributions of group behaviour approximate those of real agents?
 \item \textbf{Learning the economic transition model ($\theta^*$).} How can we learn $P_\theta(\cdot\mid\mathbf s_t,\mathbf a_t)$ so that simulated state transitions approximate those of the real economy?
 \item \textbf{Systematic evaluation of economic dynamics ($\mathcal D$).} How can we design an evaluation that identifies where simulated trajectory distributions diverge from those of the real economy? This evaluation should guide refinement of the agent policies, the economic transition model and the dynamics they jointly generate.
\end{enumerate}

A further question concerns how to represent the economic state. Many applied economic studies select state variables based on expert judgement or the variables recorded in statistical datasets. We draw on both, with economic theory guiding our choice of state variables. Whether the resulting representation captures the information needed to predict agent decisions and state transitions remains open. Developing methods to construct and evaluate more accurate state representations is an important direction for future research.

Overall, our multi-agent framework for economic world models links realistic agent policies ($\varphi^*$), an accurate transition model ($\theta^*$) and rigorous evaluation ($\mathcal D$) to improve predictions of economic dynamics. Agent policies and the economic transition model can be implemented using large language models, multilayer perceptrons (MLPs) or small transformers, with structured economic models as an additional option for economic transitions. When these predictions are reliable, comparing counterfactual trajectories under alternative policies, institutions and exogenous shocks can help human and AI decision-makers choose actions that better serve their economic objectives.

\section{Implementing an economic world model}
\label{sec:implementation}

In this section, we provide an initial implementation of EWM by learning household policies from microdata through behavioural cloning and calibrating a structured economic transition model against macroeconomic data using Sobol--RBF search. We first specify the economic model and its agents; policy learning then supplies the actions needed for transition calibration.

\subsection{A structured economic transition model}

The economic transition model maps the current economic state and joint action to the next state. Learning this mapping from scratch with a purely data-driven generative model remains difficult. Economics, however, has a long tradition of using structured models to capture how individual decisions and market interactions give rise to aggregate dynamics \citep{smets2007shocks,tesfatsion2006handbook}. We therefore construct $P_\theta$ from established economic models and refer to it as a structured economic transition model. We calibrate selected components of $\theta$ against observed trajectories so that the simulated dynamics better match their real-world counterparts.

We instantiate the structured transition model with \EconGym{}, a modular simulation environment spanning multiple economic models \citep{mi2025econgym}. \EconGym{} models households, firms, banks and governments and links their actions through budget constraints, production technology, market settlement and balance-sheet accounting. The equations below illustrate one setting with Ramsey households, a representative firm, a commercial bank and a fiscal authority. See \citet[Sec.~3 and Appendix~C]{mi2025econgym} for the underlying platform specifications.

\paragraph{Ramsey households.}
Ramsey households, modelled as infinite-horizon agents that choose consumption and saving to maximise discounted lifetime utility, are a standard benchmark for intertemporal choice and long-run growth in macroeconomics. At time $t$, household $i$ observes its assets $a_t^i$, labour efficiency $e_t^i$ and public variables such as the wage rate $W_t$. It chooses hours worked $h_t^i$ and earns labour income $W_te_t^ih_t^i$. Household $i$'s financial wealth comprises a bank balance $s_t^i$ and risky investment $v_t^i$, so $a_t^i=s_t^i+v_t^i$. Let $x^+=\max(x,0)$. Deposits $(s_t^i)^+$ earn the deposit rate $r_{t-1}^d$, whereas loans $(-s_t^i)^+$ incur the lending rate $r_{t-1}^l$. Risky investment earns the return $\rho_{t-1}$.

Household $i$'s lifetime utility over consumption and labour is
\begin{equation}
 U^i=\mathbb E_0\sum_{t=0}^{\infty}\beta^t
 \left[\frac{(c_t^i)^{1-\sigma}}{1-\sigma}
 -\frac{(h_t^i)^{1+\gamma}}{1+\gamma}\right].
 \label{eq:bridge-household-utility}
\end{equation}
Here, $\beta$ discounts future utility, $\sigma$ controls relative risk aversion and $\gamma$ is the inverse Frisch elasticity of labour supply.

Household $i$ pays taxes $T_t^i$ to the government and allocates its remaining resources between consumption $c_t^i$ at price $p_t$ and next-period assets $a_{t+1}^i$, subject to
\begin{equation}
 p_tc_t^i+a_{t+1}^i
 =(1+r_{t-1}^d)(s_t^i)^+
 -(1+r_{t-1}^l)(-s_t^i)^+
 +(1+\rho_{t-1})v_t^i
 +W_te_t^ih_t^i-T_t^i.
 \label{eq:bridge-household-budget}
\end{equation}
This Ramsey specification illustrates one household type; other household types and their transition rules are detailed in \citet[Sec.~3 and Appendix~C]{mi2025econgym}.

\paragraph{Representative firm.}
The representative firm stands in for the economy's producers, converting capital and labour into goods. It combines labour $L_t$ supplied by households with capital $K_t$ financed by the bank. With total factor productivity $Z_t$ and capital elasticity of output $\alpha$, its output $Y_t$ follows a Cobb--Douglas production function:
\begin{equation}
 Y_t=Z_tK_t^{\alpha}L_t^{1-\alpha}.
 \label{eq:bridge-production}
\end{equation}
Depending on market structure, the firm may set the goods price $p_t$. It pays the wage rate $W_t$ for labour and the capital rental rate $R_t^K$ for bank-financed capital. Its profit is sales revenue net of these costs:
\begin{equation}
 \Pi_t^{\mathrm{firm}}
 =p_tY_t-W_tL_t-R_t^KK_t.
 \label{eq:bridge-firm-profit}
\end{equation}

\paragraph{Commercial bank.}
Households deposit funds with the commercial bank or borrow from it. The bank finances firm capital $K_t$ and holds government bonds $B_t$. In \EconGym{}, commercial banks set the deposit rate $r_t^d$ and the lending rate $r_t^l$ charged on household loans. Let $D_t=\sum_i(s_t^i)^+$ and $H_t=\sum_i(-s_t^i)^+$ denote household deposits and loans, respectively. The bank earns the capital rental rate $R_t^K$ and bond rate $r_{t-1}$. With $A_t$ denoting net household deposits, the balance-sheet relation and period profit are
\begin{equation}
 \begin{aligned}
 A_t&=D_t-H_t=K_t+B_t,\\
 \Pi_t^{\mathrm{bank}}
 &=R_t^KK_t+r_{t-1}B_t
 +r_{t-1}^lH_t-r_{t-1}^dD_t.
 \end{aligned}
 \label{eq:bridge-bank-profit}
\end{equation}

\paragraph{Fiscal authority.}
The fiscal authority sets tax policy and determines government purchases $G_t$. Its income and asset tax parameters $\chi_t^T$ include a tax-level parameter $\tau_T$ and a progressivity parameter $\xi_T$. Tax revenue $T_t$ and new bonds $B_{t+1}$ finance purchases and service debt $B_t$ carried into the period. The government budget constraint is
\begin{equation}
 (1+r_{t-1})B_t+G_t=B_{t+1}+T_t.
 \label{eq:bridge-government-budget}
\end{equation}
The interest rate $r_{t-1}$ applies to debt carried into the period. Through taxes and spending, the fiscal authority can pursue economic growth, lower inequality and higher aggregate welfare, subject to its budget constraint.

\paragraph{Micro-to-macro dynamics.}
Household decisions generate aggregate consumption $C_t$, effective labour $L_t$ and net deposits $A_t$. Firm output is allocated among household consumption, government purchases and investment; investment also updates the capital stock:
\begin{equation}
 C_t=\sum_i c_t^i,\qquad
 L_t=\sum_i e_t^ih_t^i,\qquad
 A_t=\sum_i s_t^i,\qquad
 p_tY_t=p_tC_t+G_t+I_t,\qquad
 K_{t+1}=(1-\delta)K_t+I_t.
 \label{eq:bridge-market-clearing}
\end{equation}
The full set of transitions in the configured \EconGym{} implementation defines $P_\theta$. Together, these equations illustrate how individual decisions and market interactions generate the next economic state. We calibrate selected parameters so that simulated dynamics better match observed data.

\paragraph{Transition parameters to calibrate.}
Alongside economic variables updated each period, the structured model contains parameters that can be calibrated for a target question or setting. In the example above, these include the household discount factor $\beta$, preference parameters $\sigma$ and $\gamma$, and production elasticity $\alpha$; an economic system contains many others. We collect these transition model parameters in $\theta$:
\begin{equation}
 \theta=(\beta,\sigma,\gamma,\alpha,\ldots).
 \label{eq:calibratable-parameter-examples}
\end{equation}
We next describe how economic agent policies are learned and how transition model parameters are calibrated.

\subsection{Learning to simulate agent behaviour}
\label{sec:policy}

In this implementation, we focus on simulating the behaviour of a large, heterogeneous population of households. Government actions follow observed policy data. Firms operate under perfect competition: wages follow labour-market clearing, while goods prices adjust to supply and demand. Bank deposit and lending rates are initialised from observed base-year data.

We use behavioural cloning to learn household consumption and labour policies from microeconomic data \citep[Sec.~2.1]{ross2011reduction}. Within each task, households share policy parameters, while differences in their observed features lead to heterogeneous actions. We implement each household policy $\pi_\varphi^i$ using a neural network. This paper presents one implementation rather than prescribing a particular backbone; large language models or other computational models could also parameterise the policy. The central design choice is the learning objective. A policy trained only on static state--action mappings may fit observed decisions yet fail to predict how actions adjust when states change. Such dynamic responses are essential for counterfactual experiments under alternative policies and shocks. Individual responses aggregate into group behaviour and shape macroeconomic outcomes, so errors at the household level can distort counterfactual results. We therefore focus below on an objective that matches both action levels and their responses to changes in economic states.

For household $i$ observed at $t_0<t_1$, the policy input $o_{t_1}^i$ includes baseline household characteristics, selected changes between observations and the baseline action $a_{t_0}^i$. In our implementation, the policy predicts the action at $t_1$ by adding the change predicted by the network to $a_{t_0}^i$.
\begin{equation}
 \widehat a_{t_1}^{\,i}
 = \pi_\varphi^i(o_{t_1}^i)
 = a_{t_0}^i + \Delta\widehat a_{t_0:t_1}^{\,i},
 \qquad
 \Delta a_{t_0:t_1}^{\,i} = a_{t_1}^i - a_{t_0}^i.
 \label{eq:behavior-cloning-vector}
\end{equation}
Below, $\Delta\widehat a^{\,i}$ and $\Delta a^i$ denote $\Delta\widehat a_{t_0:t_1}^{\,i}$ and $\Delta a_{t_0:t_1}^i$, respectively. The observed action $a_{t_0}^i$ anchors the predicted level, so the error in $\widehat a_{t_1}^{\,i}$ equals the error in $\Delta\widehat a^{\,i}$. We therefore measure individual behaviour using the normalised action-change error
\[
 \mathcal L_{\mathrm{ind}}(\varphi)
 =\frac{\mathbb E_w\!\left[
   \left(\Delta\widehat a^{\,i}-\Delta a^i\right)^2
 \right]}
 {\operatorname{Var}_{\mathrm{train}}(\Delta a)},
\]
where $\mathbb E_w$ is the survey-weighted average over households. Normalising by the training-sample variance makes losses comparable across actions measured on different scales.

For a household group $g$, let $G_g$ denote the aggregation rule from Section~\ref{sec:formulation} applied to that group. Within each group, we compare predicted and observed distributions of action levels at $t_1$ and of action changes from $t_0$ to $t_1$. The resulting group loss is
\[
 \mathcal L_{\mathrm{group}}(\varphi)
 =\mathbb E_g\!\left[
   \frac{W_2^2\!\left(G_g(\widehat a_{t_1}),G_g(a_{t_1})\right)}
        {\operatorname{Var}_{\mathrm{train}}(a)}
   +
   \frac{W_2^2\!\left(G_g(\Delta\widehat a),G_g(\Delta a)\right)}
        {\operatorname{Var}_{\mathrm{train}}(\Delta a)}
 \right],
\]
where $\mathbb E_g$ averages over household groups and $W_2$ is the Wasserstein distance \citep{peyre2019optimal}. Combining individual and group errors gives
\begin{equation}
 \mathcal L_{\mathrm{policy}}(\varphi)
 =\mathcal L_{\mathrm{ind}}(\varphi)
 +\lambda_G\mathcal L_{\mathrm{group}}(\varphi),
 \qquad
 \widehat\varphi=\arg\min_\varphi\mathcal L_{\mathrm{policy}}(\varphi).
 \label{eq:policy-training-loss}
\end{equation}
We estimate the household policy parameters $\widehat\varphi$ from microeconomic data using the individual and group losses above. We then hold the fitted policies fixed while calibrating the economic transition model in Section~\ref{sec:transition-calibration}.

\subsection{Calibrating the economic transition model}
\label{sec:transition-calibration}

Given the trained economic agent policies parameterised by $\widehat\varphi$, the learned policies generate a joint action $\widehat{\mathbf a}_t$ from the current state $\widehat{\mathbf s}_t$. The transition model $P_\theta$ then maps this state--action pair to the next state $\widehat{\mathbf s}_{t+1}$. Repeating this process from the observed initial state $\mathbf s_0$ produces a simulated trajectory. We define the environment loss as the discrepancy between this trajectory and observed economic dynamics, and calibrate $\theta$ by solving
\begin{equation}
\begin{gathered}
 \widehat\theta
 =\arg\min_{\theta\in\Theta}
 \mathcal L_{\mathrm{env}}(\tau_{\widehat\varphi,\theta}),\\
 \text{s.t.}\qquad
 \widehat{\mathbf a}_t
 \sim\widetilde\Pi_{\widehat\varphi}
 (\cdot\mid\widehat{\mathbf s}_t),
 \qquad
 \widehat{\mathbf s}_{t+1}
 \sim P_\theta
 (\cdot\mid\widehat{\mathbf s}_t,\widehat{\mathbf a}_t),
 \qquad t=0,\ldots,T-1,\\
 \tau_{\widehat\varphi,\theta}
 =(\mathbf s_0,\widehat{\mathbf a}_0,\widehat{\mathbf s}_1,\ldots,
 \widehat{\mathbf a}_{T-1},\widehat{\mathbf s}_T),
 \qquad
 \widehat{\mathbf s}_0\equiv\mathbf s_0.
\end{gathered}
 \label{eq:transition-calibration-objective}
\end{equation}
Equation~\eqref{eq:transition-calibration-objective} defines the rollout that generates the simulated trajectory $\tau_{\widehat\varphi,\theta}$. We calibrate the economic transition model using observed macroeconomic data. Because complete real-world counterparts are unavailable for many elements of the simulated trajectory, $\mathcal L_{\mathrm{env}}$ compares the simulated and observed paths of available macroeconomic variables, including output, aggregate consumption, productive capital, debt, aggregate labour and wages.

Let $\mathbf x_t$ and $\widehat{\mathbf x}_t(\theta)$ denote the observed and simulated $d$-dimensional economic-variable vectors at time $t\in\mathcal T$, where $\mathcal T$ is the calibration window. From an economic perspective, we use period-to-period log growth to characterise local dynamics and log levels relative to a base period $t_0$ to capture accumulated path deviations. We define the corresponding feature maps as
\begin{equation}
 g_t(\mathbf x_t,\mathbf x_{t-1})
 :=100(\log\mathbf x_t-\log\mathbf x_{t-1}),
 \qquad
 \ell_t(\mathbf x_t;\mathbf x_{t_0})
 :=100(\log\mathbf x_t-\log\mathbf x_{t_0}).
 \label{eq:calibration-growth-level}
\end{equation}
Applying the same maps to the observed and simulated inputs gives $\mathbf f_t$ and $\widehat{\mathbf f}_t(\theta)$, respectively, for $f\in\{g,\ell\}$. Because the variables differ in volatility, raw feature errors are not directly comparable. For variable $j$, let $\sigma_{f,j}=\max\{\operatorname{SD}_{u\in\mathcal T}(f_{u,j}),\epsilon\}$, where $\epsilon>0$ prevents unstable normalisation. We define the standardised error for each feature as
\begin{equation}
 E_f(\theta)
 =\left[
  \frac{1}{d|\mathcal T|}
  \sum_{t\in\mathcal T}\sum_{j=1}^{d}
  \left(
   \frac{\widehat f_{t,j}(\theta)-f_{t,j}}
        {\sigma_{f,j}}
  \right)^2
 \right]^{1/2},
 \qquad f\in\{g,\ell\}.
 \label{eq:calibration-standardized-errors}
\end{equation}
We then combine the growth and level feature errors into the environment loss:
\begin{equation}
 \mathcal L_{\mathrm{env}}(\tau_{\widehat\varphi,\theta})
 =\left[
  \lambda E_g^{2}(\theta)
  +(1-\lambda)E_\ell^{2}(\theta)
 \right]^{1/2}
 \label{eq:environment-calibration-loss}
\end{equation}
Here, $\lambda\in[0,1]$ controls the trade-off between matching period-to-period dynamics and the accumulated level path. Equation~\eqref{eq:environment-calibration-loss} assigns each candidate $\theta$ a single calibration loss, and minimising it yields $\widehat\theta$ in equation~\eqref{eq:transition-calibration-objective}. We next present a practical parameter-search method for solving equation~\eqref{eq:transition-calibration-objective}.

\paragraph{Sobol--RBF parameter search.}
Each loss evaluation requires a multi-period simulation of interacting agent decisions and economic transitions, so gradients with respect to $\theta$ are not readily available. We therefore introduce a derivative-free optimisation method for calibrating $\theta$. Sobol sampling \citep{sobol1967distribution} explores the bounded parameter space, while a radial basis function (RBF) surrogate \citep{regis2007stochastic} fitted to evaluated parameter--loss pairs guides subsequent evaluations.

Let $p$ be the number of selected transition parameters. We rescale each parameter from its search bounds to $[0,1]$, so a candidate is $\mathbf u\in[0,1]^p$. An initial Sobol sample explores this space. For each evaluated candidate $\mathbf u_b$, we recover $\theta_b$, simulate an economic trajectory and compute $L_b=\mathcal L_{\mathrm{env}}(\tau_{\widehat\varphi,\theta_b})$ using equation~\eqref{eq:environment-calibration-loss}. After $n$ evaluations, we fit an RBF surrogate to these parameter--loss pairs:

\begin{equation}
 \widetilde{\mathcal L}_n(\mathbf u)
 =\sum_{b=1}^{n}\omega_b\,
 \rho(\lVert\mathbf u-\mathbf u_b\rVert_2)
 +c_0+\mathbf c^{\top}\mathbf u,
 \label{eq:rbf-surrogate}
\end{equation}
Here, $\rho$ is a thin-plate spline basis, and $\omega_b$, $c_0$ and $\mathbf c$ are coefficients fitted to the evaluated losses with smoothing. The surrogate estimates losses for untested candidates and guides subsequent simulations.

At iteration $n$, a Sobol candidate pool $\mathcal U_n$ is ranked by predicted loss and distance $D_n(\mathbf u)=\min_{b\leq n}\lVert\mathbf u-\mathbf u_b\rVert_2$ from the nearest evaluated point:
\begin{equation}
 A_n(\mathbf u)
 =\widetilde{\mathcal L}_n(\mathbf u)
 -\gamma_n D_n(\mathbf u),
 \qquad
 \mathbf u_{n+1}
 =\arg\min_{\mathbf u\in\mathcal U_n}A_n(\mathbf u),
 \label{eq:rbf-acquisition}
\end{equation}
The weight $\gamma_n\geq0$ balances low predicted loss against exploration. We map the selected $\mathbf u_{n+1}$ back to $\theta_{n+1}$ using the parameter bounds, run a full simulation to obtain $L_{n+1}=\mathcal L_{\mathrm{env}}(\tau_{\widehat\varphi,\theta_{n+1}})$, and add $(\mathbf u_{n+1},L_{n+1})$ to the pairs used to refit the surrogate. If a stable fit is unavailable, we instead evaluate a distant Sobol candidate.

After $B$ candidate evaluations, we select the lowest-loss parameter vector from the set $\mathcal C_B$ of fully evaluated candidates:
\begin{equation}
 \widehat\theta
 =\arg\min_{\theta\in\mathcal C_B}
 \mathcal L_{\mathrm{env}}(\tau_{\widehat\varphi,\theta}).
 \label{eq:calibrated-environment-parameters}
\end{equation}

Algorithm~\ref{alg:ewm-implementation} summarizes how this section provides an initial implementation of the general framework introduced in Section~\ref{sec:formulation}. It links microeconomic and macroeconomic data by using the former to train agent policies and the latter to calibrate the economic transition model. Coupling the learned policies with the calibrated transition model connects individual actions to aggregate economic dynamics, which are evaluated in Section~\ref{sec:evaluation}.

\begin{ewmalgorithmbox}{Implementing an Economic World Model}
\label{alg:ewm-implementation}
\textbf{Input:} Microeconomic and macroeconomic datasets; initial state $\mathbf s_0$; horizon $T$.

\textbf{Output:} Fitted policy parameters $\widehat\varphi$; calibrated transition parameters $\widehat\theta$; simulated dynamics $\tau_{\widehat\varphi,\widehat\theta}$.

\begin{ewmalgorithmsteps}
 \algstep \textbf{Transition model:} Specify $P_\theta$ and its calibration domain $\Theta$.
 \algstep \textbf{Policy learning:} Fit $\widehat\varphi\gets\argmin_\varphi\mathcal L_{\mathrm{policy}}(\varphi)$ to the microeconomic data.
 \algstep \textbf{Transition calibration:} With $\widehat\varphi$ fixed, solve $\widehat\theta\gets\argmin_{\theta\in\Theta}\mathcal L_{\mathrm{env}}(\tau_{\widehat\varphi,\theta})$ against the macroeconomic data using Sobol--RBF search.
 \algstep \textbf{Rollout:} Set $\widehat{\mathbf s}_0\gets\mathbf s_0$.
 \algstep \textbf{for} $t=0,\ldots,T-1$ \textbf{do}
 \algstep \hspace*{1em}\textbf{for each} agent $i\in\mathcal I_t$ \textbf{do}
 \algstep \hspace*{2em}Receive observation $\widehat o_t^i\sim O_i(\cdot\mid\widehat{\mathbf s}_t)$ and take action $\widehat a_t^i\sim\pi_{\widehat\varphi}^{\,i}(\cdot\mid\widehat o_t^i)$.
 \algstep \hspace*{1em}\textbf{end for}
 \algstep \hspace*{1em}Collect the active agents' actions into the joint action $\widehat{\mathbf a}_t\gets (\widehat a_t^i)_i$.
 \algstep \hspace*{1em}Transition to the next state $\widehat{\mathbf s}_{t+1}\sim P_{\widehat\theta}(\cdot\mid\widehat{\mathbf s}_t,\widehat{\mathbf a}_t)$.
 \algstep \textbf{end for}
 \algstep \textbf{return Economic Dynamics:} $\tau_{\widehat\varphi,\widehat\theta}=(\mathbf s_0,\widehat{\mathbf a}_0,\widehat{\mathbf s}_1,\ldots,\widehat{\mathbf a}_{T-1},\widehat{\mathbf s}_T)$.
\end{ewmalgorithmsteps}
\end{ewmalgorithmbox}

\Needspace{.56\textheight}
\section{A micro-to-macro evaluation framework for economic world models}
\label{sec:evaluation}

\begin{figure}[H]
\centering
\includegraphics[width=\textwidth]{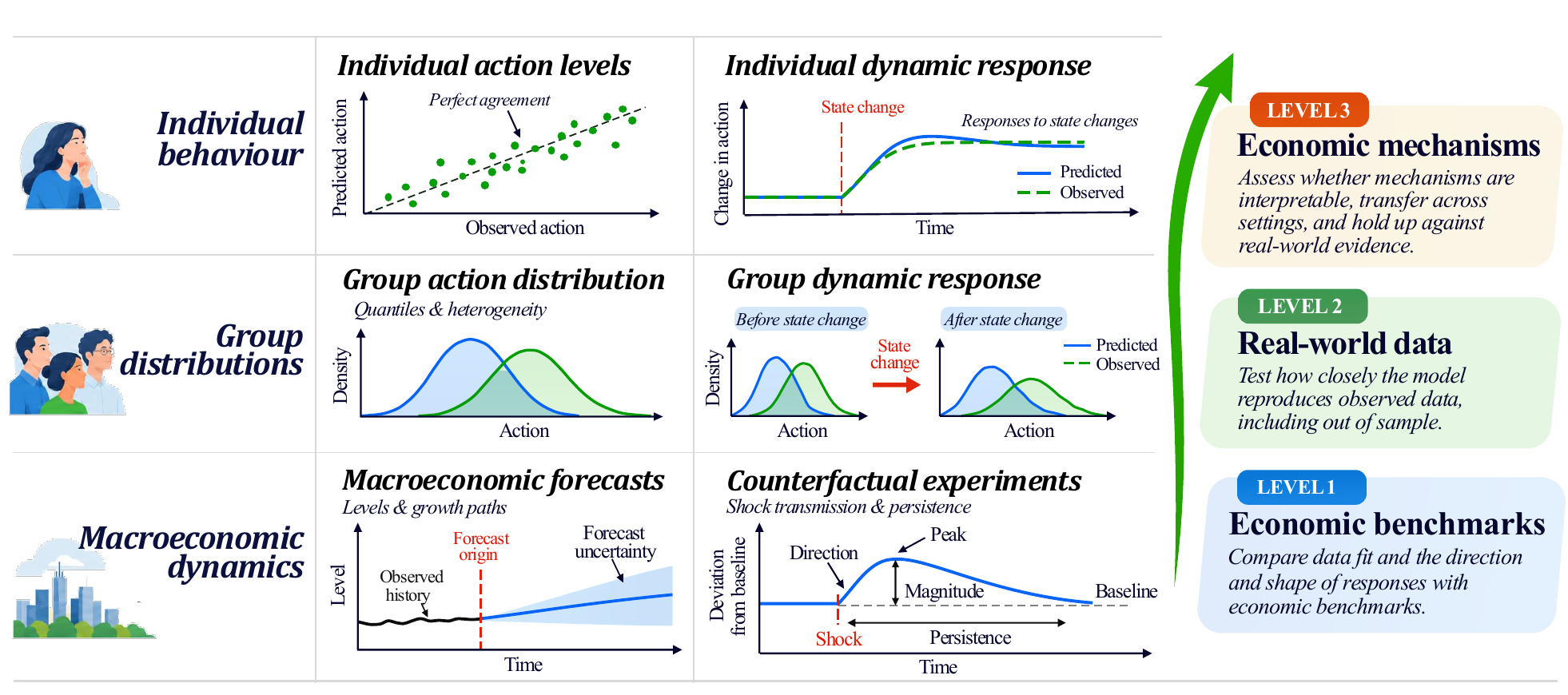}
\caption{\textbf{Micro-to-macro evaluation framework for economic world models.} We assess individual action levels and responses, group action distributions and their shifts, and macroeconomic dynamics through forecasts and counterfactuals. We compare performance with economic benchmarks, test fit to real-world data including held-out observations, and assess whether mechanisms are interpretable, transferable and supported by evidence.}
\label{fig:evaluation-framework}
\end{figure}

Sections~\ref{sec:formulation} and~\ref{sec:implementation} show how interactions among agents and economic transitions generate economic dynamics. This section proposes a systematic evaluation framework $\mathcal D$ for economic world models (Figure~\ref{fig:evaluation-framework}), spanning micro to macro: individual action levels and responses, group behaviour distributions and their dynamic responses, and macroeconomic dynamics. At the macroeconomic level, we evaluate forecasts and counterfactual experiments against real-world data and economic benchmarks, and examine whether the resulting patterns and mechanisms are economically interpretable.

Figure~\ref{fig:evaluation-framework} (left) shows how we evaluate an EWM from individual behaviour to macroeconomic dynamics. The right panel defines three capability levels, each setting a higher standard for EWM development.

\noindent\textcircled{\scriptsize 1}\enspace\textbf{Level~1. Matching economic benchmark models.} An EWM performs at least as well as an economic benchmark model on the task at hand. The benchmark reflects what economists can currently explain or predict about the problem.

\noindent\textcircled{\scriptsize 2}\enspace\textbf{Level~2. Reproducing real-world data.} An EWM reproduces key patterns in real-world data and the economic phenomena those patterns reflect.

\noindent\textcircled{\scriptsize 3}\enspace\textbf{Level~3. Explaining economic mechanisms.} Different economic mechanisms can produce similar outcomes. Identifying the mechanisms at work therefore requires evidence beyond fit to observed data. An EWM at this level captures interpretable mechanisms that are consistent with established economic findings and hold across relevant settings. More capable EWMs can also generate testable hypotheses about previously unrecognised mechanisms. When supported by independent evidence, these hypotheses can inform economic analysis and policy decisions.

\subsection{Evaluating individual behaviour}

We assess whether agent policies reproduce observed actions and how agents adjust their actions as the economic state changes.

\paragraph{Individual action levels.}
For each action, such as consumption or labour, let $a_{t_1}^i$ and $\widehat a_{t_1}^{\,i}$ denote agent $i$'s observed and predicted actions at $t_1$, as defined in equation~\eqref{eq:behavior-cloning-vector}. The operator $\mathbb E_w$ denotes a survey-weighted mean over the observations entering each metric, with equal weights when unavailable. We measure their discrepancy using the weighted root mean squared error (RMSE):
\begin{equation}
 \operatorname{RMSE}_{\mathrm{level}}
 =\left[\mathbb E_w\!\left(
 (\widehat a_{t_1}^{\,i}-a_{t_1}^i)^2
 \right)\right]^{1/2},\qquad
 \operatorname{NRMSE}_{\mathrm{level}}
 =\frac{\operatorname{RMSE}_{\mathrm{level}}}{\sigma_a}.
 \label{eq:individual-level-rmse}
\end{equation}
Here, $\sigma_a$ is the weighted standard deviation of observed action levels in the evaluation sample. RMSE is expressed in the action's units, whereas normalised RMSE (NRMSE) scales the error by the observed variation of that action.

\paragraph{Individual dynamic responses.}
For matched observations of the same agent at $t_0<t_1$, we evaluate the observed and predicted action changes $\Delta a^i$ and $\Delta\widehat a^{\,i}$ defined in equation~\eqref{eq:behavior-cloning-vector}. The two observations need not be consecutive, but the interval should remain short enough to limit shifts in the underlying decision rule. Each pair uses its survey weight at $t_1$:
\begin{equation}
\begin{aligned}
 \operatorname{RMSE}_{\Delta}
 &=\left[\mathbb E_w\!\left(
 (\Delta\widehat a^{\,i}-\Delta a^i)^2
 \right)\right]^{1/2},
 &\qquad
 \operatorname{NRMSE}_{\Delta}
 &=\frac{\operatorname{RMSE}_{\Delta}}{\sigma_{\Delta a}}.
\end{aligned}
 \label{eq:micro-metrics}
\end{equation}
Here, $\sigma_{\Delta a}$ is the weighted standard deviation of observed action changes in the evaluation sample. We evaluate changes over different time spans separately. Because each CEX prediction is anchored to the observed initial action, level and change RMSE share the same residual; their NRMSEs differ because action levels and changes have different observed variances.

\subsection{Evaluating group distributions}

Household microdata provide a limited, noisy view of real-world behaviour. A policy that closely fits each observed action may also capture incidental variation, increasing the risk of overfitting. We therefore assess action distributions within household groups and how they shift as economic states change.

\paragraph{Group action distributions.}
We assess whether the policy reproduces the distribution of actions within groups of agents. For each action, let $\mathcal I_{g,t}$ denote the agents in group $g$ at time $t$. We define groups using only characteristics observed before the action. Applying the aggregation rule $G_g$ from Section~\ref{sec:policy} to the observed and predicted actions gives
\begin{equation*}
 \mu_{g,t}=G_g\!\left(\{a_t^i\}_{i\in\mathcal I_{g,t}}\right),\qquad
 \widehat\mu_{g,t}=G_g\!\left(\{\widehat a_t^{\,i}\}_{i\in\mathcal I_{g,t}}\right).
\end{equation*}
The observed and predicted distributions use the same agents and survey weights, with equal weights when survey weights are unavailable. A group may also contain the full population.

\textbf{Distributional distance} measures how far predicted and observed action distributions differ across quantiles. We use the 2-Wasserstein distance ($W_2$) \citep{peyre2019optimal}. Writing $\widehat\mu_{g,t}$ and $\mu_{g,t}$ as $\widehat\mu$ and $\mu$ for a given group and period, we calculate
\begin{equation}
 W_2^2(\widehat\mu,\mu)
 =\int_0^1|Q_{\widehat\mu}(q)-Q_\mu(q)|^2\,\mathrm dq.
 \label{eq:group-metrics}
\end{equation}
Here, $Q_\mu(q)$ is the weighted $q$th quantile of observed actions, and $Q_{\widehat\mu}(q)$ is its predicted counterpart. We normalise $W_2$ by the relevant training-sample standard deviation when comparing actions with different units.

\textbf{Distributional overlap} measures the probability mass shared by the predicted and observed distributions within the same action ranges. For common action bins $\mathcal B$, we calculate the overlap coefficient (OVL) as
\begin{equation}
 \operatorname{OVL}(\widehat\mu,\mu)
 =\sum_{B\in\mathcal B}\min\{\widehat\mu(B),\mu(B)\}.
 \label{eq:group-overlap}
\end{equation}
For each bin $B$, the shared mass is the smaller of the predicted and observed survey-weighted shares. OVL sums this shared mass across bins and ranges from zero to one, with one indicating complete overlap at the chosen resolution.

\paragraph{Group dynamic responses.}
We assess whether the policy reproduces how action distributions shift as agents' economic states change. For selected times $t_0<t_1$, let $\mathcal I_g(t_0,t_1)$ contain agents with comparable changes in information observed before the later decision, such as a decline or rise in income. Applying $G_g$ to the action changes in equation~\eqref{eq:behavior-cloning-vector} gives
\begin{equation*}
 \mu_g^{\Delta}=G_g\!\left(\{\Delta a^i\}_{i\in\mathcal I_g(t_0,t_1)}\right),\qquad
 \widehat\mu_g^{\Delta}=G_g\!\left(\{\Delta\widehat a^{\,i}\}_{i\in\mathcal I_g(t_0,t_1)}\right).
\end{equation*}
Both distributions contain the same agents and use the same survey weights. We assess their distance and overlap using the metrics above:
\begin{equation}
 W_{2,\Delta}(g;t_0,t_1)=W_2(\widehat\mu_g^{\Delta},\mu_g^{\Delta}),\qquad
 \operatorname{OVL}_{\Delta}(g;t_0,t_1)
 =\operatorname{OVL}(\widehat\mu_g^{\Delta},\mu_g^{\Delta}).
 \label{eq:group-dynamic-distance}
\end{equation}
These comparisons assess the direction, magnitude and heterogeneity of group responses, including changes in dispersion and tails. They describe conditional behaviour and do not by themselves identify causal effects of state changes.

\subsection{Evaluating economic dynamics}

Individual actions and group action distributions can be compared directly with observed microdata. Economic transitions are not directly observed; they are abstractions distilled from real-world economic phenomena and activities. We therefore evaluate simulated economic dynamics through two complementary tasks. Macroeconomic forecasting tests whether the model reproduces observed paths, whereas counterfactual experiments test whether responses to interventions exhibit coherent economic mechanisms.

\subsubsection{Macroeconomic forecasting}

We assess aggregate paths with fixed-origin and rolling-origin forecasts and score the EWM and our implementation of the Gertler model against observed data and the reference proxies described in Section~\ref{sec:experiment-setting}.

\paragraph{Indicators and error measures.}
Our forecasting experiments evaluate real output, consumption, productive capital, older-group wealth share, capital return and government debt relative to gross domestic product (GDP). For each positive indicator, we reuse the annual log-growth and log-level error measures in equation~\eqref{eq:calibration-growth-level}. We denote either transformed series by $z_t^m$, where $m\in\{\mathrm{growth},\mathrm{level}\}$. The common base-period term is omitted from the log level because it cancels from forecast errors. We report RMSE separately for each indicator and transformation.

\paragraph{Fixed-origin forecasting.}
Fixed-origin forecasting starts from one observed state at origin $t_0$ and generates an uninterrupted path for at most $H$ periods without later observed states. Here, $h$ denotes the number of periods ahead, and $\mathcal H_{\mathrm{obs}}\subseteq\{1,\ldots,H\}$ contains the horizons with observed outcomes. The root mean squared error (RMSE) is
\begin{equation}
 \operatorname{RMSE}^{\mathrm{fix}}_m
 =\left[
 \frac{1}{|\mathcal H_{\mathrm{obs}}|}
 \sum_{h\in\mathcal H_{\mathrm{obs}}}
 \left(
  \widehat z_{t_0+h\mid t_0}^{m}-z_{t_0+h}^{m}
 \right)^2
 \right]^{1/2}.
 \label{eq:fixed-rmse}
\end{equation}
Because the model is not reinitialised, fixed-origin forecasts also reveal how errors accumulate over the forecast horizon.

\Needspace{7\baselineskip}
\paragraph{Rolling-origin forecasting.}
Rolling-origin forecasting advances the origin through a sequence of historical dates. At each origin $t$, we reinitialise the model from the observed state and forecast up to $H$ periods ahead. For a given horizon $h\leq H$, $\mathcal T_h$ contains the origins whose target at $t+h$ is observed. The horizon-specific RMSE is
\begin{equation}
 \operatorname{RMSE}^{\mathrm{roll}}_m(h)
 =\left[
 \frac{1}{|\mathcal T_h|}
 \sum_{t\in\mathcal T_h}
 \left(
  \widehat z_{t+h\mid t}^{m}-z_{t+h}^{m}
 \right)^2
 \right]^{1/2}.
 \label{eq:rolling-rmse}
\end{equation}
We report RMSE separately by horizon because both the number and composition of available origins vary with $h$. At each horizon, we compare models using the same origin--target pairs.

\subsubsection{Counterfactual experiments}

Impulse-response analysis and scenario analysis are widely used to study macroeconomic shocks and policy interventions \citep{sims1980,smets2007shocks,delNegroSchorfheide2013}. Counterfactuals are generally constructed in two ways. The first exogenously imposes a path for a policy-relevant variable. The second introduces an exogenous shock, often scaled to one standard deviation. We then compare the resulting counterfactual path with a matched baseline, holding the initial state, agent policies, transition parameters and random seed fixed. This comparison probes mechanisms that historical fit alone cannot establish \citep{lucas1976econometric}.

For intervention $s$, let $\widehat X_{s,h}$ and $\widehat X_{0,h}$ denote the simulated values under the intervention and matched baseline paths. We define their percentage deviation from baseline as
\begin{equation}
 D_s(h)
 =100\left(\frac{\widehat X_{s,h}}{\widehat X_{0,h}}-1\right).
 \label{eq:counterfactual-relative-response}
\end{equation}
For shares and ratios stored as fractions, we report the percentage-point deviation $D_s^{\mathrm{pp}}(h)=100(\widehat X_{s,h}-\widehat X_{0,h})$. A zero or near-zero baseline requires a level difference or another prespecified scale. With multiple seeds, we calculate each intervention path's deviation from its matched baseline before summarising across seeds.

The resulting deviation paths reveal response direction, peak magnitude and timing, persistence, terminal effects and transmission across variables. Benchmark comparisons align the intervention, initial state, evaluation window and response measure wherever possible. Counterfactual responses have no observed ground truth. We therefore assess structural plausibility by comparing their direction, shape, timing and cross-variable transmission with economic benchmarks and by examining whether the implied mechanisms are interpretable. These comparisons do not identify real-world causal effects.

This micro-to-macro evaluation framework quantifies errors in observed individual, group and aggregate outcomes and assesses the structural plausibility of counterfactual dynamics. Together, these diagnostics provide feedback for agent policy training and economic transition model calibration.

\section{Experiments}
\label{sec:experiments}

\renewcommand{\topfraction}{0.95}
\renewcommand{\bottomfraction}{0.90}
\renewcommand{\textfraction}{0.05}
\renewcommand{\floatpagefraction}{0.80}
\setcounter{topnumber}{4}
\setcounter{bottomnumber}{2}
\setcounter{totalnumber}{6}
\setlength{\textfloatsep}{10pt plus 2pt minus 2pt}
\setlength{\floatsep}{8pt plus 2pt minus 2pt}
\setlength{\intextsep}{8pt plus 2pt minus 2pt}
\captionsetup[figure]{skip=5pt}
\captionsetup[table]{skip=4pt}
\makeatletter
\setlength{\@fptop}{0pt}
\setlength{\@fpsep}{8pt}
\setlength{\@fpbot}{0pt plus 1fil}
\makeatother

Following the evaluation framework in Section~\ref{sec:evaluation}, we assess EWM across individual actions, group behaviour distributions and macroeconomic dynamics. We then test whether it helps LLM agents make better policy decisions. Four research questions guide the experiments:
\begin{itemize}[leftmargin=2em,itemsep=.2em]
 \item \textbf{RQ1:} How closely do household policies match observed actions and group distributions? (\S~\ref{sec:rq1})
 \item \textbf{RQ2:} How accurately does EWM forecast macroeconomic dynamics? (\S~\ref{sec:rq2})
 \item \textbf{RQ3:} What do counterfactual experiments reveal about EWM's economic responses and mechanisms? (\S~\ref{sec:rq3})
 \item \textbf{RQ4:} Can EWM help LLM agents make better policy decisions? (\S~\ref{sec:rq4})
\end{itemize}

\subsection{Experimental setting}
\label{sec:experiment-setting}

We use US microeconomic data to train and evaluate policies for a large, heterogeneous household population, and US macroeconomic data to calibrate the transition model and assess EWM's forecasts of economic dynamics. Our case study examines government borrowing in a life-cycle economy, comparing EWM with the benchmark model of \citet{gertler1999government}. Future work will examine the framework's generalisability across countries and economic questions.

\paragraph{Microeconomic data.}
Household policies use survey-weighted records from the 2014--2024 Consumer Expenditure Survey (CEX) Interview Survey \citep{blsCEX2024}. Repeated quarterly interviews track households over time, allowing us to train policies to predict how their actions change with economic conditions. We train both policies on 2014--2018 records, select checkpoints using 2019 validation records and test them on 2020--2024 records (Table~\ref{tab:micro-data-split}).

\begin{table}[!htbp]
\centering\small
\caption{CEX household-policy samples by action-label year. For example, 32,401 (19,371) denotes 32,401 eligible interview pairs from 19,371 distinct households. Some households contribute multiple pairs.}
\label{tab:micro-data-split}
\begin{tabular}{L{2.7cm}C{3.6cm}C{3.2cm}C{3.6cm}}
\toprule
Task & Training (2014--2018) & Validation (2019) & Test (2020--2024)\\
\midrule
Consumption & 32,401 (19,371) & 6,741 (4,612) & 22,347 (14,920)\\
Labour & 14,607 (14,607) & 3,146 (3,146) & 13,399 (13,399)\\
\bottomrule
\end{tabular}
\end{table}

\paragraph{Macroeconomic data.}
The macroeconomic experiments use annual US aggregate data for 2014--2025. Output, consumption, hours, compensation, prices and federal debt come from Federal Reserve Economic Data (FRED) \citep{fredData2026}; productive capital comes from the Bureau of Economic Analysis (BEA) Fixed Assets Accounts \citep{beaFixedAssets2026,beaNationalAccounts2026}; and age-group wealth shares come from the Federal Reserve's Distributional Financial Accounts \citep{federalReserveDFA2026}. We construct a capital-return proxy from the capital--output ratio and model parameters, and a real Treasury-return proxy from the one-year Treasury rate and a price index \citep{beaNationalAccounts2026,federalReserveGS1,fredData2026}. The fixed-origin experiment starts from the 2014 state, calibrates transition parameters against 2015--2019 outcomes, and evaluates forecasts out of time in 2020--2025. Rolling-origin forecasts update their starting state and recalibrate transition parameters using data available through each origin (Section~\ref{sec:rq2}).

\paragraph{Economic questions and benchmark model.}
We examine how government borrowing affects household consumption and saving, capital accumulation and wealth across workers and retirees. Gertler's overlapping-generations (OLG) model provides a benchmark for this question, tracing the short- and long-run effects of government debt and social security in a life-cycle economy \citep{gertler1999government}. Its later use as the basis for a New Keynesian life-cycle model shows how this structure has informed subsequent research \citep{fujiwara2008lifecycle}.

\Needspace{5\baselineskip}
\subsection{RQ1: How closely do household policies match observed actions and group distributions?}
\label{sec:rq1}

We train household consumption and labour policies on 2014--2018 records, select checkpoints using 2019 validation records, and evaluate them on 2020--2024 test records. We assess how closely the learned policies match observed individual action levels and changes, group action distributions, and changes in those distributions.

\subsubsection{Individual action levels and dynamic responses}

\begin{table}[!htbp]
\centering\small
\caption{Individual household action errors by sample. NRMSE is normalised within each sample (Section~\ref{sec:evaluation}); lower is better.}
\label{tab:policy-individual-results}
\begin{tabular*}{\textwidth}{@{\extracolsep{\fill}}llccc@{}}
\toprule
Action & Outcome & Training (2014--2018) & Validation (2019) & Test (2020--2024)\\
\midrule
Consumption & Individual action level & 0.5464 & 0.5280 & 0.5458\\
Consumption & Individual response & 0.9231 & 0.8564 & 0.8828\\
Labour & Individual action level & 0.5224 & 0.4822 & 0.5209\\
Labour & Individual response & 0.8808 & 0.9665 & 0.9415\\
\bottomrule
\end{tabular*}
\end{table}

\begin{figure}[!htbp]
\centering
\includegraphics[width=\textwidth]{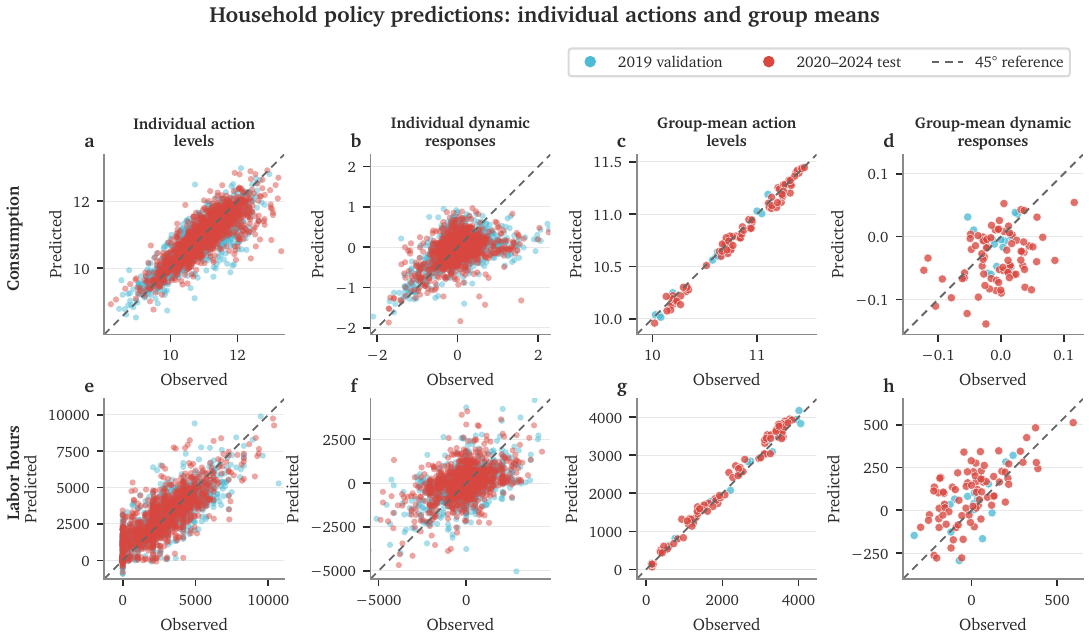}
\caption{Observed versus predicted household actions and group means. Semi-transparent filled circles show a fixed random subsample of individual records; group means use all year, income and age cells. Points nearer the dashed 45-degree line indicate smaller errors. Cyan denotes the 2019 validation sample and red the 2020--2024 temporal test; all metrics use the full samples.}
\label{fig:policy-oos}
\end{figure}

\textbf{The learned policies reproduced the main patterns of individual consumption and labour levels.} Predicted actions cluster around the 45-degree line in both dimensions (Figure~\ref{fig:policy-oos}), retaining much of the cross-household variation in observed actions. Test level NRMSEs were 0.546 for consumption and 0.521 for labour (Table~\ref{tab:policy-individual-results}). These endpoint predictions combine observed starting actions with predicted changes.

\textbf{The policies also captured patterns in how individual actions respond to changing states.} The response panels show a positive observed--predicted pattern, but their points are more dispersed around the 45-degree line than those for action levels (Figure~\ref{fig:policy-oos}). Test response NRMSEs were 0.883 for consumption and 0.941 for labour (Table~\ref{tab:policy-individual-results}). These scores provide a reference for improving response accuracy.

\subsubsection{Group distributions and dynamic responses}

\begin{table}[!htbp]
\centering\small
\caption{Distributional accuracy of household action levels and changes in the 2020--2024 test sample. Group $W_2$ is aggregated over year $\times$ training-income tertile $\times$ age-band cells ($<35$, 35--49, 50--64, $\geq65$); OVL pools test records. $\sigma_{\rm train}$ is the corresponding training-sample standard deviation (Section~\ref{sec:evaluation}).}
\label{tab:policy-group-results}
\begin{tabular*}{\textwidth}{@{\extracolsep{\fill}}llccc@{}}
\toprule
Action & Outcome & Group $W_2$ $\downarrow$ & Group $W_2/\sigma_{\rm train}$ $\downarrow$ & Overall OVL (\%) $\uparrow$\\
\midrule
Consumption & Group distribution level & 0.147 & 0.195 & 94.9\\
Consumption & Group distribution response & 0.159 & 0.355 & 90.9\\
Labour & Group distribution level & 462 & 0.236 & 78.0\\
Labour & Group distribution response & 477 & 0.411 & 73.3\\
\bottomrule
\end{tabular*}
\end{table}

\begin{figure}[!htbp]
\centering
\includegraphics[width=\textwidth]{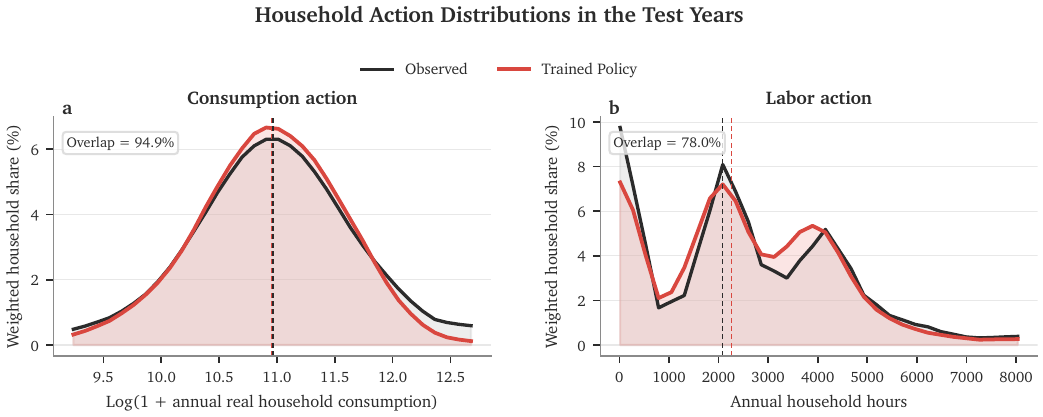}
\vspace{4mm}
\includegraphics[width=\textwidth]{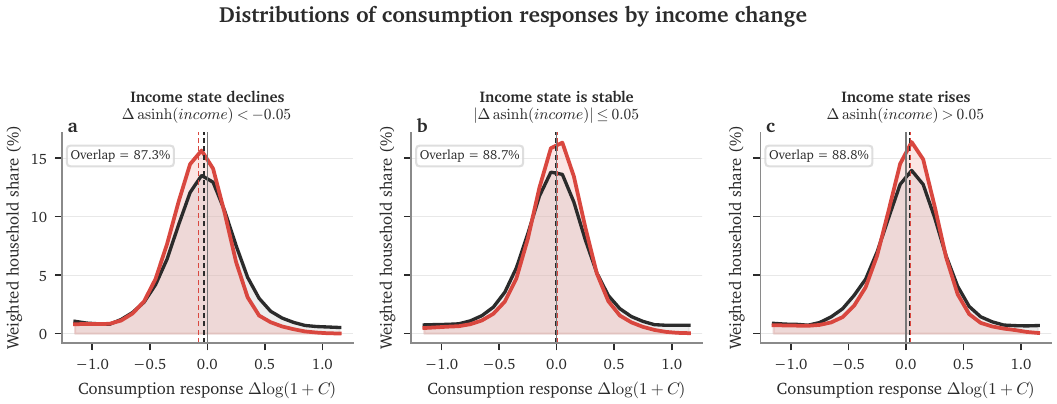}
\caption{Observed and predicted household action distributions in 2020--2024. Upper panels show consumption levels and labour hours; lower panels show consumption changes by observed pre-decision income change (9,870 of 22,347 records). Black denotes observations, red raw predictions, dashed lines weighted medians, and panel labels report OVL. Consumption $C$ is annual real household consumption in 2018 US dollars.}
\label{fig:policy-group-distributions}
\end{figure}

\textbf{Group action-level distributions showed high overlap with observed data.} The overlap coefficient (OVL) was 94.9\% for consumption and 78.0\% for labour (Table~\ref{tab:policy-group-results}; Figure~\ref{fig:policy-group-distributions}, upper panels). Labour predictions captured positive-hours modes but underrepresented nonworkers: 24.1\% reported zero hours, while only 7.0\% of raw predictions were negative and would be projected to zero. Aggregation can temper individual noise and expose this systematic shortfall.

\textbf{Group response distributions followed observed income-linked shifts and revealed local magnitude differences.} Pooled response OVL was 90.9\% for consumption and 73.3\% for labour (Table~\ref{tab:policy-group-results}). Across declining, stable and rising income, conditional consumption OVL was 87.3\%, 88.7\% and 88.8\% (Figure~\ref{fig:policy-group-distributions}, lower panels). Predicted centres aligned when income was stable or rose. When income declined, the predicted median consumption decrease was slightly larger ($-0.078$ predicted versus $-0.032$ observed).

\subsection{RQ2: How accurately does EWM forecast macroeconomic dynamics?}
\label{sec:rq2}

Having evaluated individual and group behaviour, we calibrate the economic transition model and couple it with the learned household policies to form a complete EWM. We then test how well it reproduces observed macroeconomic dynamics and forecasts subsequent years.

\subsubsection{Calibration of the economic transition model}

\begin{figure}[H]
\centering
\includegraphics[width=.75\textwidth]{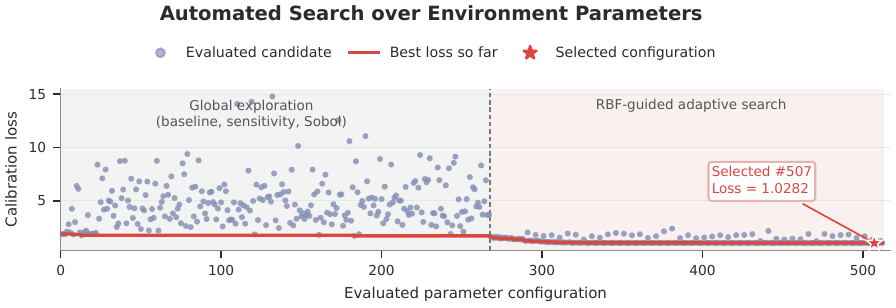}
\caption{Sobol--RBF search for environment parameters. Points show losses from fully simulated parameter combinations; the red line tracks the lowest evaluated loss, and the star marks the selected combination. Shading separates initial exploration from RBF-guided search.}
\label{fig:calibration-search}
\end{figure}

With the trained household policies fixed, we initialise the simulation in 2014 and calibrate the transition parameters against observed macroeconomic outcomes from 2015 to 2019. We evaluated each candidate on three stochastic paths using the Sobol--RBF procedure in Section~\ref{sec:transition-calibration}; the growth and level-path loss in equation~\eqref{eq:environment-calibration-loss} guided selection (Figure~\ref{fig:calibration-search}). Table~\ref{tab:calibration-loss} compares the initial and selected parameters on these search paths, five new paths over the same years, and those new paths in 2020--2025. The latter two evaluations were performed after selection.

\begin{table}[!htbp]
\centering\small
\caption{Mean macroeconomic path loss under the initial scenario and calibrated parameters. The search used three stochastic paths; independent historical and test evaluations used five new paths. Lower is better.}
\label{tab:calibration-loss}
\begin{tabular}{L{6.4cm}C{2.3cm}C{2.3cm}C{2.3cm}}
\toprule
Years and purpose & Initial parameters & Calibrated parameters & Relative improvement\\
\midrule
2015--2019, search paths & 1.8727 & 1.0282 & 45.1\%\\
2015--2019, paths excluded from search & 1.9601 & 1.3137 & 33.0\%\\
2020--2025, test-period paths & 4.4457 & 3.7004 & 16.8\%\\
\bottomrule
\end{tabular}
\end{table}

\FloatBarrier

\textbf{RBF-guided search found the lowest evaluated loss.} Following 267 baseline, sensitivity and Sobol evaluations, we simulated 245 RBF-guided candidates. The best observed loss fell from 1.6978 to 1.0282 at candidate 507. Relative to the initial scenario, losses fell by 45.1\% on search paths, 33.0\% on new historical paths and 16.8\% in test years (Table~\ref{tab:calibration-loss}). The initial scenario included historical fiscal settings; the selected parameters also passed eight ten-year stability simulations.

\subsubsection{Fixed-origin multistep forecasting}

We initialise the fixed-origin EWM rollout with the 2014 state and use parameters calibrated on 2015--2019 outcomes. We hold those parameters fixed and roll forward recursively through 2025. The test period begins in 2020. The archived Gertler benchmark ends in 2024, so the paired error comparison covers 2020--2024.

\begin{figure}[!htbp]
\centering
\includegraphics[width=\textwidth]{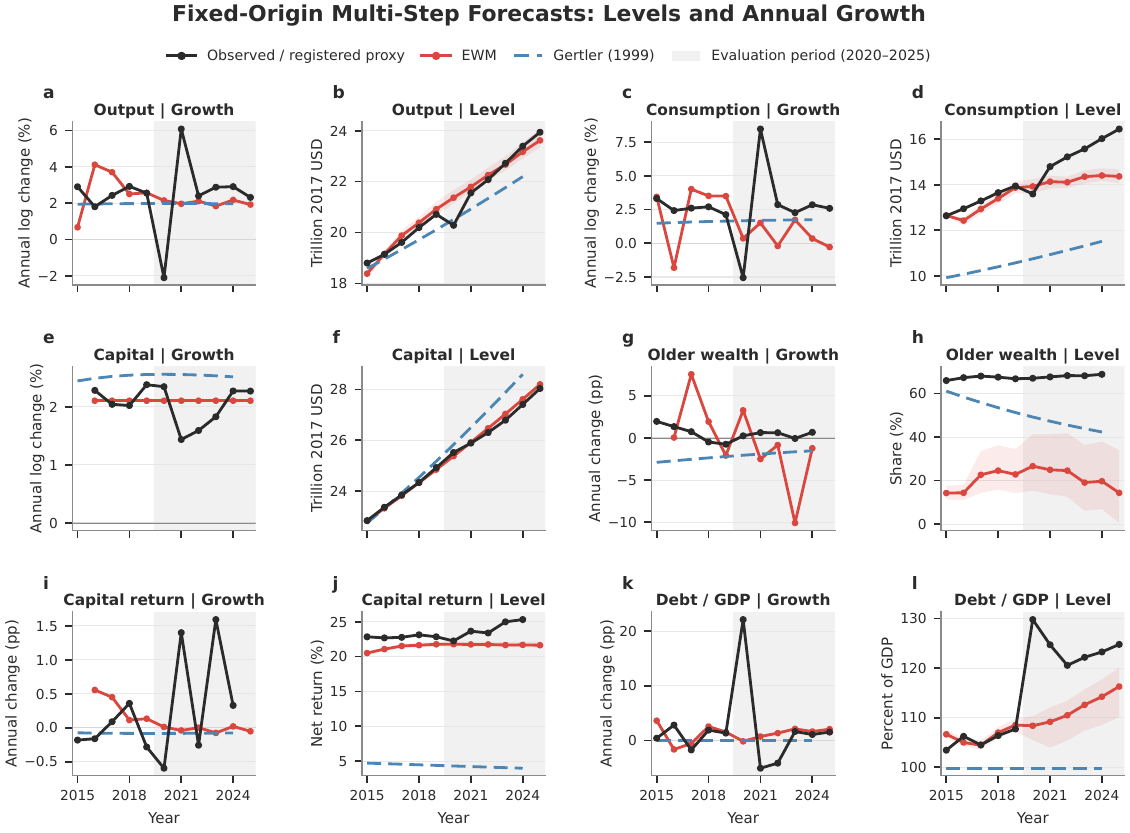}
\caption{Fixed-origin forecasts of output, consumption, capital, older-group wealth share, capital return and debt/GDP. Output, consumption and capital levels are in trillions of chained 2017 US dollars; other levels are percentages. Growth panels report annual log changes for real aggregates and annual percentage-point changes for shares and rates. Black denotes observations or registered proxies, red EWM and blue our implementation of the Gertler model. Red bands show the 10th--90th percentiles of EWM level paths.}
\label{fig:fixed-forecast}
\end{figure}

\begin{table}[!htbp]
\centering\small
\caption{Fixed-origin forecast RMSE (2020--2024) for EWM and the Gertler benchmark. Log-growth errors are in percentage points; log-level errors are $100$ times the forecast-to-observed log ratio.}
\label{tab:fixed-forecast-errors}
\begin{tabular}{lcccc}
\toprule
& \multicolumn{2}{c}{Log-growth RMSE (pp) $\downarrow$} & \multicolumn{2}{c}{Log-level RMSE ($\times100$) $\downarrow$}\\
\cmidrule(lr){2-3}\cmidrule(lr){4-5}
Indicator & EWM & Benchmark & EWM & Benchmark\\
\midrule
Output & 2.71 & 2.66 & \textbf{2.45} & 3.69\\
Consumption & 3.82 & 3.65 & \textbf{7.27} & 30.23\\
Capital & \textbf{0.42} & 0.74 & \textbf{0.65} & 3.21\\
Older-group wealth share & 20.63 & 4.46 & 110.16 & 40.46\\
Capital return & \textbf{0.84} & 0.85 & \textbf{2.01} & 17.41\\
Debt/GDP & 8.92 & 8.70 & \textbf{11.87} & 21.93\\
\bottomrule
\end{tabular}
\end{table}

\textbf{EWM reproduced several macroeconomic level paths, especially output and productive capital.} Over 2020--2024, both followed observed trends, with log-level RMSEs of 2.45 and 0.65 (Figure~\ref{fig:fixed-forecast}; Table~\ref{tab:fixed-forecast-errors}). Consumption, capital return and debt/GDP levels were also closer to their reference series than the benchmark paths: their EWM errors were 7.27, 2.01 and 11.87, versus 30.23, 17.41 and 21.93. EWM thus had lower level errors for the five indicators scored against common reference series. The return comparison uses a constructed proxy. Older-group wealth shares use different definitions and are shown descriptively.

\textbf{Annual growth and abrupt turning points remain harder to forecast.} EWM's capital-growth RMSE was lower than the benchmark's (0.42 versus 0.74; Table~\ref{tab:fixed-forecast-errors}). The models had similar errors for output and capital return, whereas EWM's errors were slightly higher for consumption and debt/GDP. Neither rollout captured the pandemic-related 2020 output contraction (Figure~\ref{fig:fixed-forecast}). Improving growth-rate and turning-point forecasts remains a target for future work.

\begin{figure}[H]
\centering
\includegraphics[width=\textwidth]{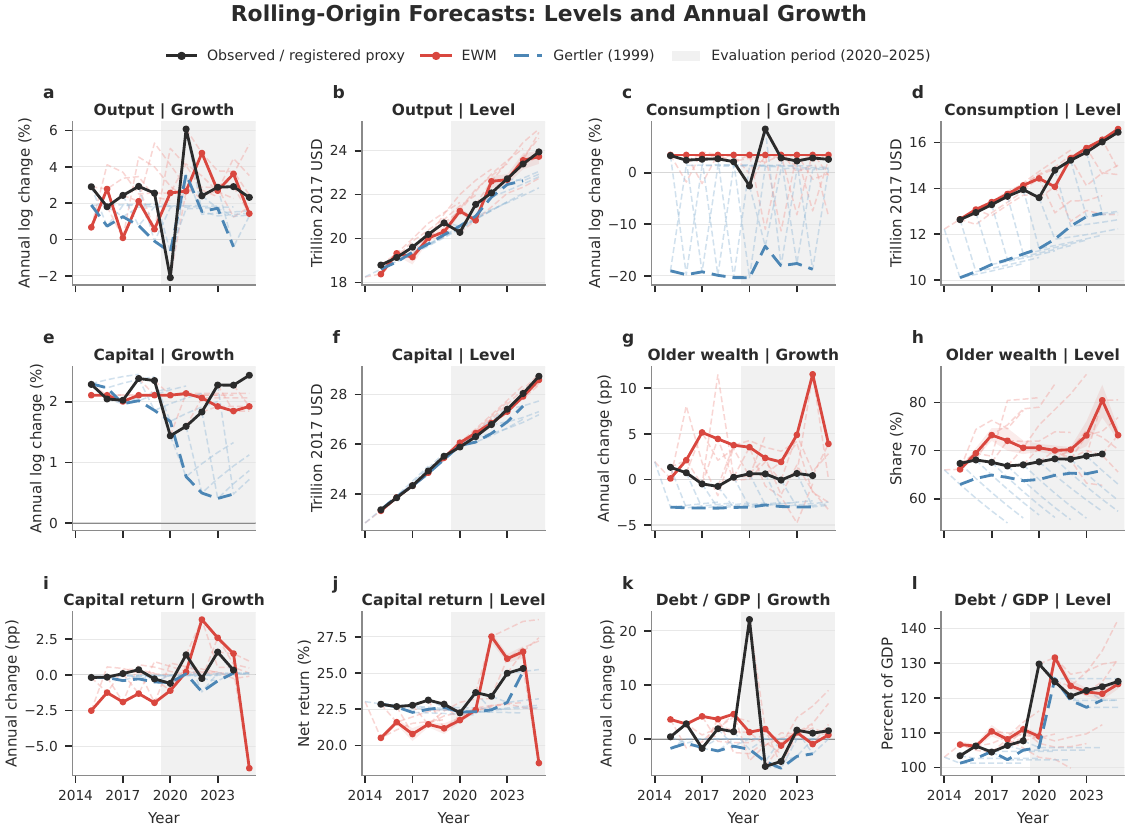}
\caption{Rolling-origin forecasts of the same macroeconomic indicators and units as Figure~\ref{fig:fixed-forecast}. Black denotes observations, red EWM and blue Gertler (1999); solid lines show one-year forecasts and pale dashed lines longer paths. Red bands show the 10th--90th percentiles of EWM one-year level forecasts across five runs.}
\label{fig:rolling-forecast}
\end{figure}

\subsubsection{Rolling-origin forecasting}

With household policies fixed, we generate five-year forecasts from each observed annual state in 2014--2024. From 2020 onward, transition parameters are recalibrated on the preceding five years; evaluation covers observed outcomes in 2020--2025.

\textbf{With annual updates to the observed state, EWM’s forecasts closely tracked observed output, consumption and capital levels, as well as the broad trends in capital returns and debt-to-GDP.} In 2024, the one-year forecasts for output, consumption and capital levels were all within 0.8\% of their observed values (Figure~\ref{fig:rolling-forecast}). The abrupt turning point caused by the pandemic was harder to anticipate. EWM first used the observed 2020 state to forecast 2021, which helps explain the lag (see Section~6.3.4 for the model’s response when the pandemic shock is imposed directly).

\begin{figure}[H]
\centering
\includegraphics[width=\textwidth]{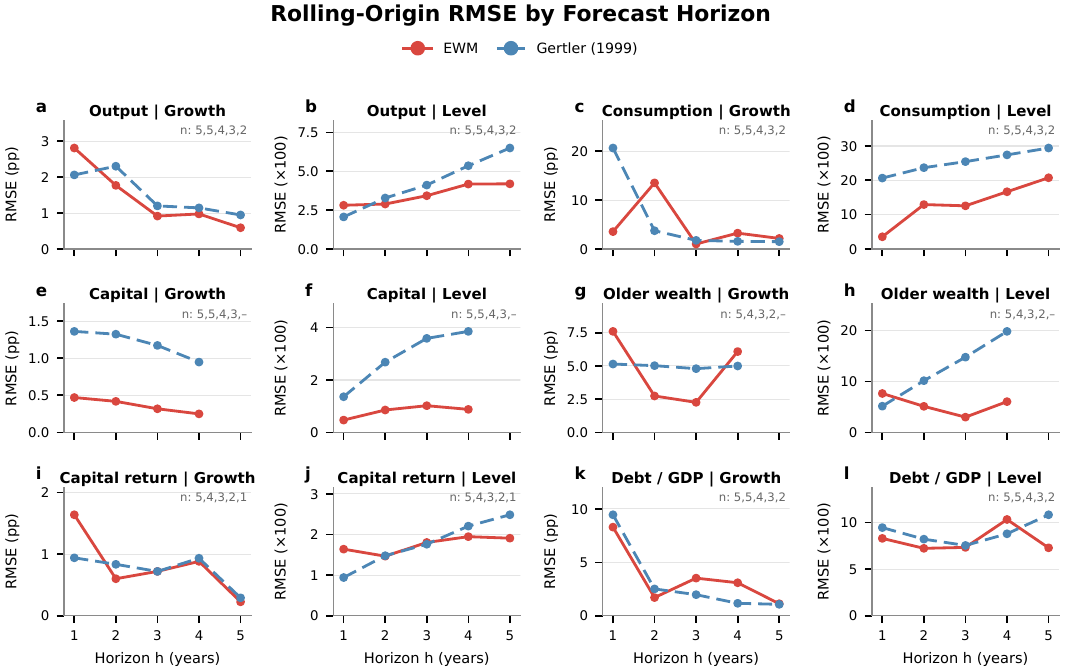}
\caption{Rolling-origin RMSE by forecast horizon for annual log growth (percentage points) and log-level error ($100\log(\widehat X/X)$). Grey $n$ gives the number of paired forecast origins; a dash in $n$ indicates no paired forecasts. Wealth shares use different groups (ages 65+ in EWM, retired households in Gertler) and cannot be compared directly.}
\label{fig:rolling-horizon-rmse}
\end{figure}

\textbf{EWM achieved lower RMSE than Gertler (1999) for most directly comparable macroeconomic forecasts.} Figure~\ref{fig:rolling-horizon-rmse} shows that EWM’s rolling-origin RMSE curves lie below Gertler’s across many variables and forecast horizons. The advantage was most consistent for capital, with lower RMSEs for both growth and levels at every comparable horizon. Consumption-level RMSEs were also lower at every comparable horizon, and RMSEs for both output growth and levels were lower from the two-year horizon onward.

\FloatBarrier
\subsubsection{Conditional simulation of the 2020 pandemic shock}

The preceding forecasts did not anticipate the abrupt 2020 pandemic shock. Once detected, we incorporate the shock into forecasts from 2020 onward and test whether EWM captures its immediate and subsequent economic effects. In 2020, realised hours are set to 96\%, 92\% or 88\% of planned hours, with policies, transition parameters and other exogenous paths fixed. We compare annual log growth with observations. These prescribed hour reductions are hypothetical, not estimates of the actual 2020 labour shock.

\begin{figure}[!t]
\centering
\includegraphics[width=\textwidth]{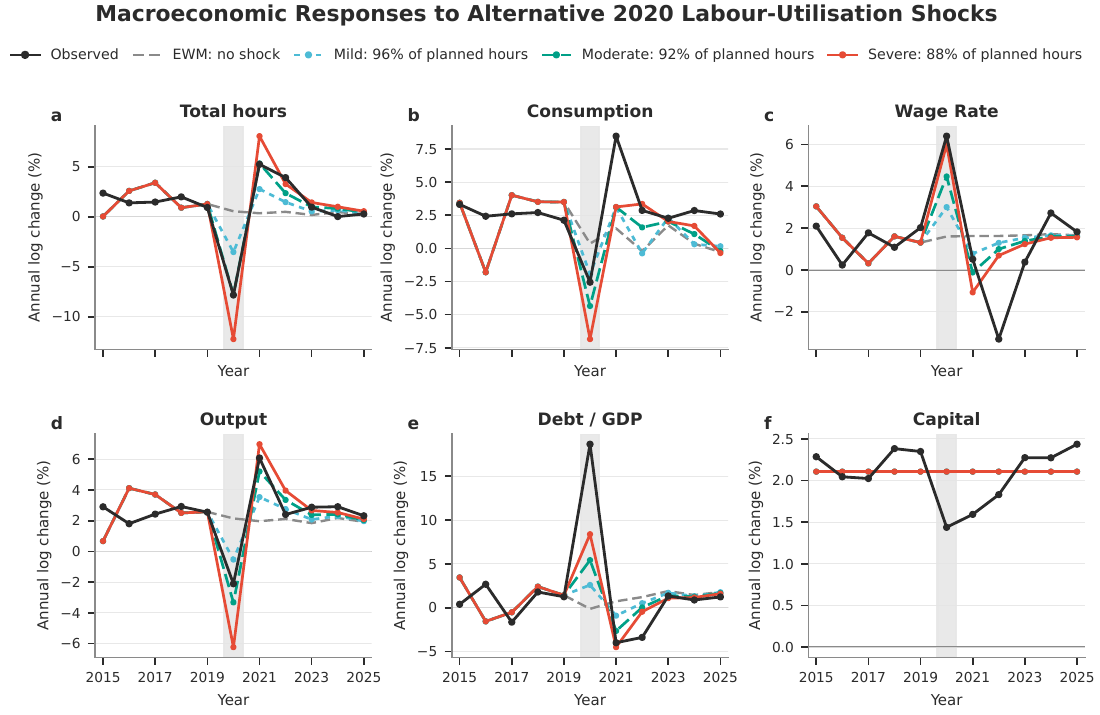}
\caption{Annual log growth in total hours, consumption, the real wage rate, output, debt/GDP and capital under alternative 2020 labour-utilisation shocks. Grey shading marks the shock year.}
\label{fig:pandemic}
\end{figure}

\textbf{Under the imposed 2020 labour shock, EWM reproduced the observed directions of change in five macroeconomic indicators.} In the 92\% hours scenario, total hours and consumption fell by $7.79\%$ and $4.35\%$, the real wage rate rose by $4.47\%$, output fell by $3.32\%$, and debt/GDP rose by $5.43\%$ (Figure~\ref{fig:pandemic}). Within the model, fewer realised hours reduce labour income, leading to lower consumption. Labour becomes scarcer relative to capital, raising the real wage rate, while lower effective labour reduces output. The resulting fall in output raises debt/GDP mainly by shrinking its denominator; the debt stock changes little.

\Needspace{6\baselineskip}
\textbf{EWM generated a 2021 rebound without a separately imposed positive shock.} After the one-year hours constraint expired, the learned household labour and consumption responses, together with the economic transition, produced growth of $5.33\%$ in hours, $3.11\%$ in consumption and $5.20\%$ in output (observed $5.24\%$, $8.47\%$ and $6.08\%$). The model thus captured the observed rebound in direction and closely matched hours and output growth, although consumption recovered less than in the data.

\FloatBarrier
\subsection{RQ3: What do counterfactual experiments reveal about EWM's economic responses and mechanisms?}
\label{sec:rq3}

\subsubsection{Counterfactual design}

\begin{figure}[H]
\centering
\includegraphics[width=.82\textwidth]{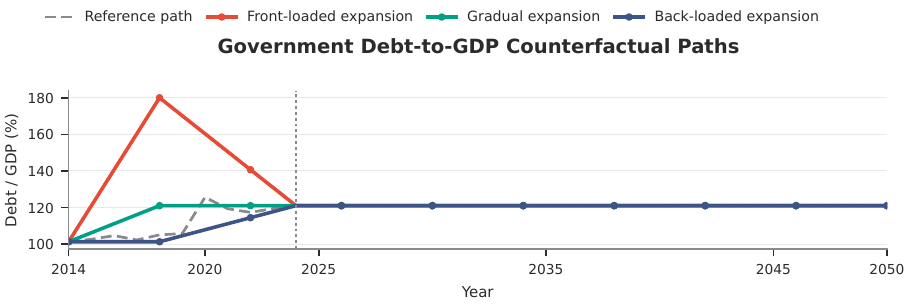}
\caption{Reference and counterfactual government-debt/GDP paths. The three borrowing paths reach a common endpoint in 2024 and remain constant thereafter.}
\label{fig:debt-paths}
\end{figure}

To examine what EWM counterfactuals reveal about economic outcomes and mechanisms, we compare three government-borrowing scenarios with a reference path. All scenarios start from the same 2014 state (debt/GDP: 101.28\%) and differ in the timing of debt accumulation. The three expansion paths converge to a debt/GDP ratio of 121.01\% in 2024 and remain constant thereafter (Table~\ref{tab:debt-paths}; Figure~\ref{fig:debt-paths}). This endpoint is anchored to the observed US gross federal debt/GDP ratio in the fourth quarter of 2024 \citep{fredDebtRatio2026}.

We compare EWM with the benchmark model of \citet{gertler1999government} under matched debt paths. In EWM, equal per-household net taxes adjust annually to implement each path, while other fiscal settings, agent policies and transition parameters remain fixed. We run five simulations per path through 2050 to assess response magnitude, timing and persistence.

\begin{table}[H]
\centering\small
\caption{Key points on the counterfactual government-debt paths (\% of GDP).}
\label{tab:debt-paths}
\begin{tabular}{L{4.4cm}C{2.5cm}C{2.5cm}C{2.5cm}}
\toprule
Scenario & 2014 & 2018 & 2024 onwards\\
\midrule
Reference path & 101.28 & 105.05 & 121.01\\
Front-loaded expansion & 101.28 & 180.00 & 121.01\\
Gradual expansion & 101.28 & 121.01 & 121.01\\
Back-loaded expansion & 101.28 & 101.28 & 121.01\\
\bottomrule
\end{tabular}
\end{table}

\subsubsection{Dynamic responses through the tax channel}

\begin{figure}[!t]
\centering
\includegraphics[width=.90\textwidth]{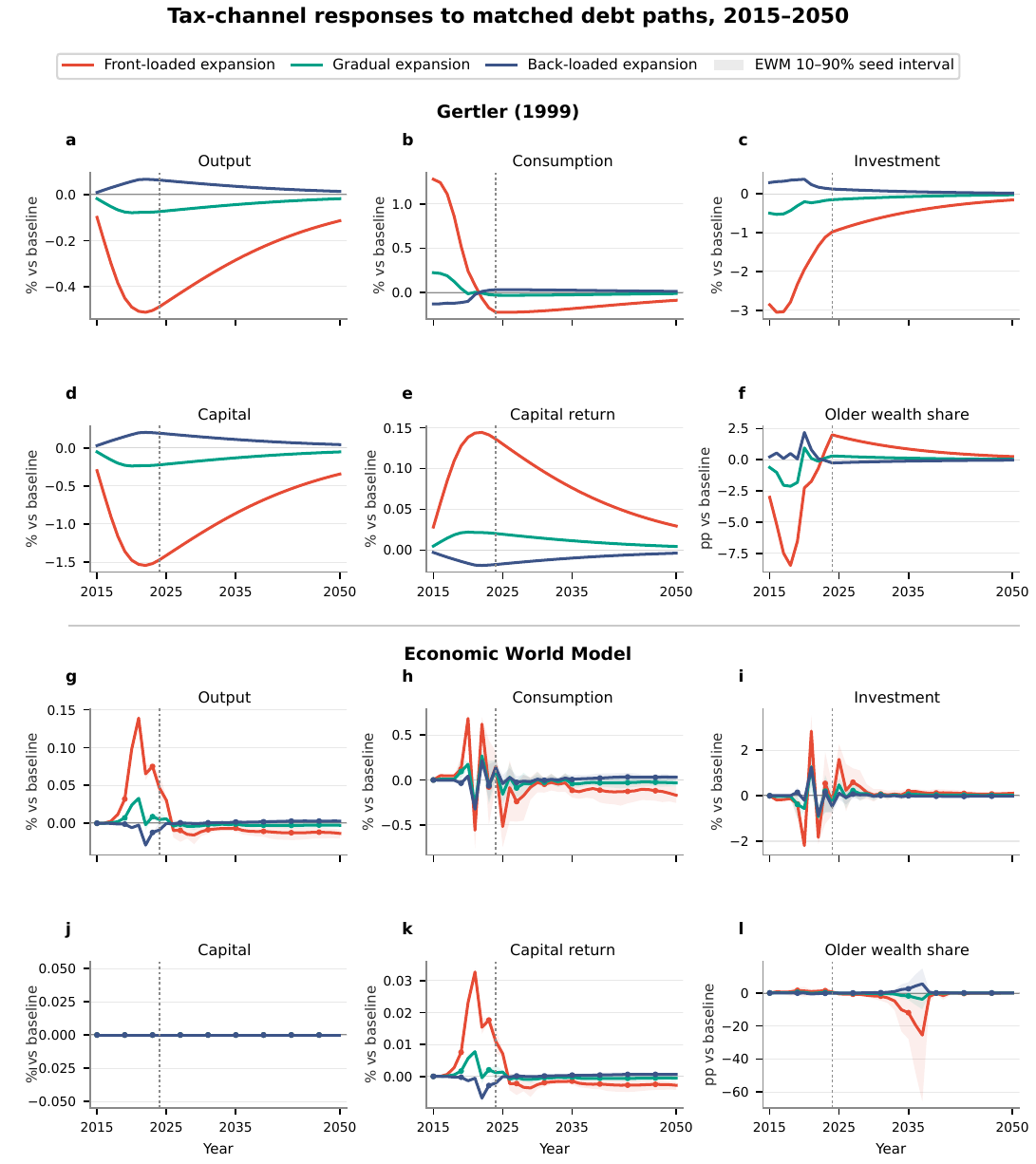}
\caption{Tax-channel responses to the three debt paths relative to the reference path, 2015--2050. Gertler (1999) is shown above EWM; colours match Figure~\ref{fig:debt-paths}. EWM lines show means across five paired seeds, with 10th--90th percentile bands.}
\label{fig:debt-response-current-2050}
\end{figure}

\textbf{EWM showed a short-run response similar to the life-cycle model of \citet{gertler1999government}: front-loaded borrowing raised consumption and capital returns but lowered total investment.} In Gertler's model, borrowing postpones part of the tax burden and stimulates current consumption demand. Relative to each model's reference path, consumption rose by $0.684\%$ in EWM and $0.242\%$ in Gertler, while total investment fell by $2.193\%$ and $1.948\%$, respectively. Capital returns rose by $0.023\%$ and $0.138\%$, respectively (Figure~\ref{fig:debt-response-current-2050}). Within EWM, larger early debt expansions produced larger 2020 responses in consumption and total investment, forming a clear short-run gradient.

\textbf{EWM's lower total investment did not translate into the capital decline seen in Gertler.} Under front-loaded borrowing, EWM's fixed investment matched its reference path in every seed through 2020; the fall in total investment came from inventory investment. EWM's productive capital remained unchanged and output rose by $0.098\%$. In Gertler, productive capital fell by $1.475\%$ and output by $0.489\%$. Capital returns thus rose for different reasons: output increased at unchanged capital in EWM, while capital fell proportionally more than output in Gertler. EWM also did not reproduce Gertler's change in the older group's wealth share, although the groups differ (ages 65 and over versus retirees).

\textbf{Overall, EWM captured part of Gertler's short-run borrowing response but did not reproduce its transmission to capital and output.} Because these counterfactuals have no observed ground truth, the comparison cannot establish which model is correct.

\FloatBarrier
\subsection{RQ4: Can EWM help LLM agents make better policy decisions?}
\label{sec:rq4}

To test whether an Economic World Model (EWM) improves LLM policy decisions, we compare direct LLM decision-making (Direct LLM) with EWM-guided LLM policy planning (LLM w. EWM). Under LLM w. EWM, the LLM agent proposes candidate policies, simulates their potential economic consequences with EWM, and selects a policy based on those simulations. Both methods pursue GDP in tax and pension policy tasks, with and without a prespecified epidemic shock. We also evaluate tax policies aimed at maximizing social welfare and pension policies that adjust retirement ages and contribution rates to increase pension-fund accumulation, our proxy for sustainability. The prespecified epidemic shock reduces realised hours to 88\% of planned hours and lowers log productivity by 0.06 in 2020; both shock components are removed in 2021. We use DeepSeek Flash\footnote{DeepSeek-V4.1-Flash-1017.} in these experiments; future work will evaluate additional LLMs.

\subsubsection{EWM-guided planning for GDP}

\begin{table}[H]
\centering\small
\caption{Cumulative rewards under the GDP objective. Values are means across five paired seeds; paired gain averages each seed's percentage increase over the direct LLM.}
\label{tab:rq4-performance}
\setlength{\tabcolsep}{4pt}
\begin{tabular}{llrrr}
\toprule
Department & Scenario & LLM & LLM w. EWM & Paired gain\\
\midrule
Tax & No shock & 6.4769 & 6.8706 & +6.13\%\\
Pension & No shock & 6.8095 & 6.8937 & +1.24\%\\
Tax & Epidemic shock & 6.3711 & 6.7879 & +6.55\%\\
Pension & Epidemic shock & 6.7146 & 6.8290 & +1.71\%\\
\bottomrule
\end{tabular}
\end{table}

\begin{figure}[!t]
\centering
\includegraphics[width=\textwidth]{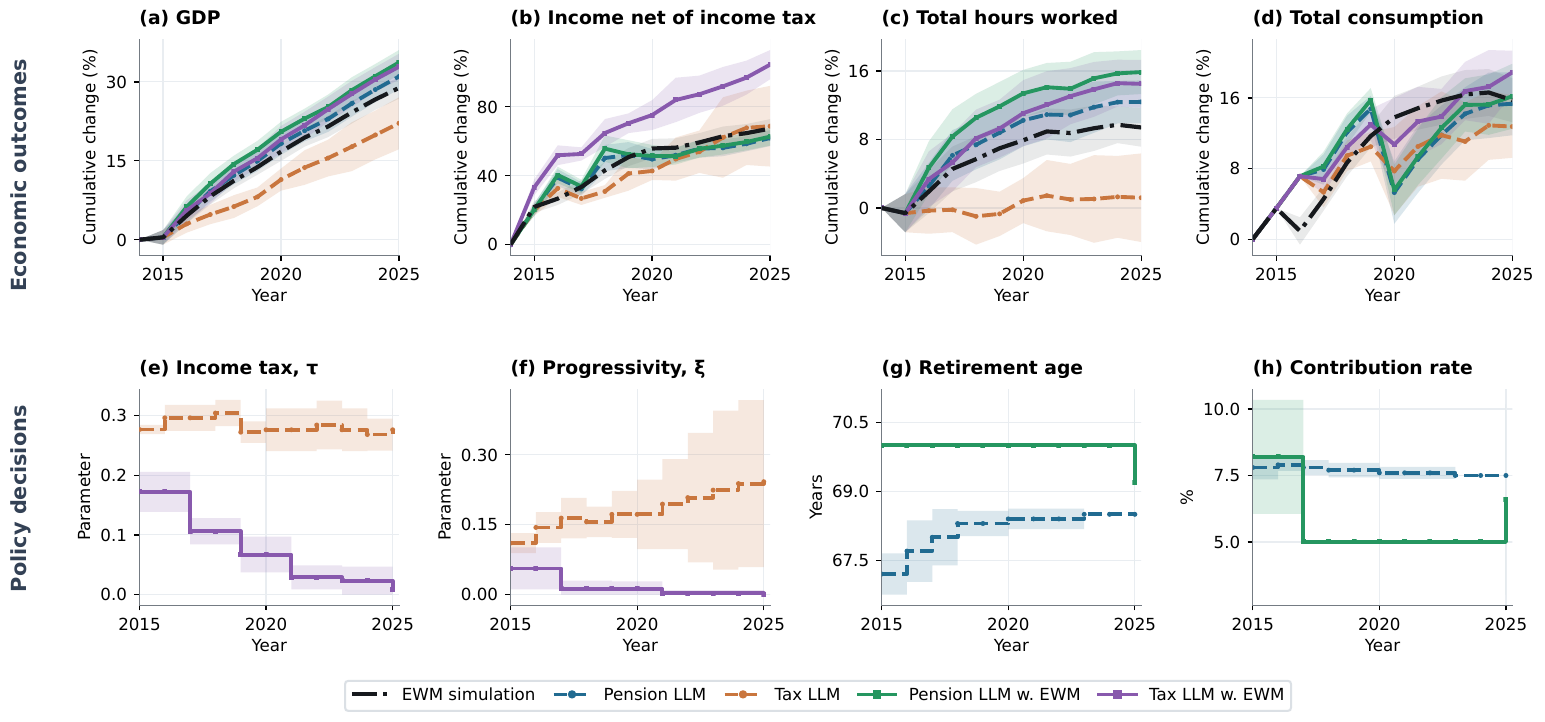}
\caption{Tax and pension policy tasks under the GDP objective without the epidemic shock. Upper panels report cumulative percentage changes from 2014 in GDP, income net of income tax, total hours worked and total consumption; lower panels report the corresponding tax and pension policy decisions. Curves show seed means with one-standard-deviation bands; black lines denote historical-policy EWM simulations.}
\label{fig:rq4-no-shock}
\end{figure}

\begin{figure}[!t]
\centering
\includegraphics[width=\textwidth]{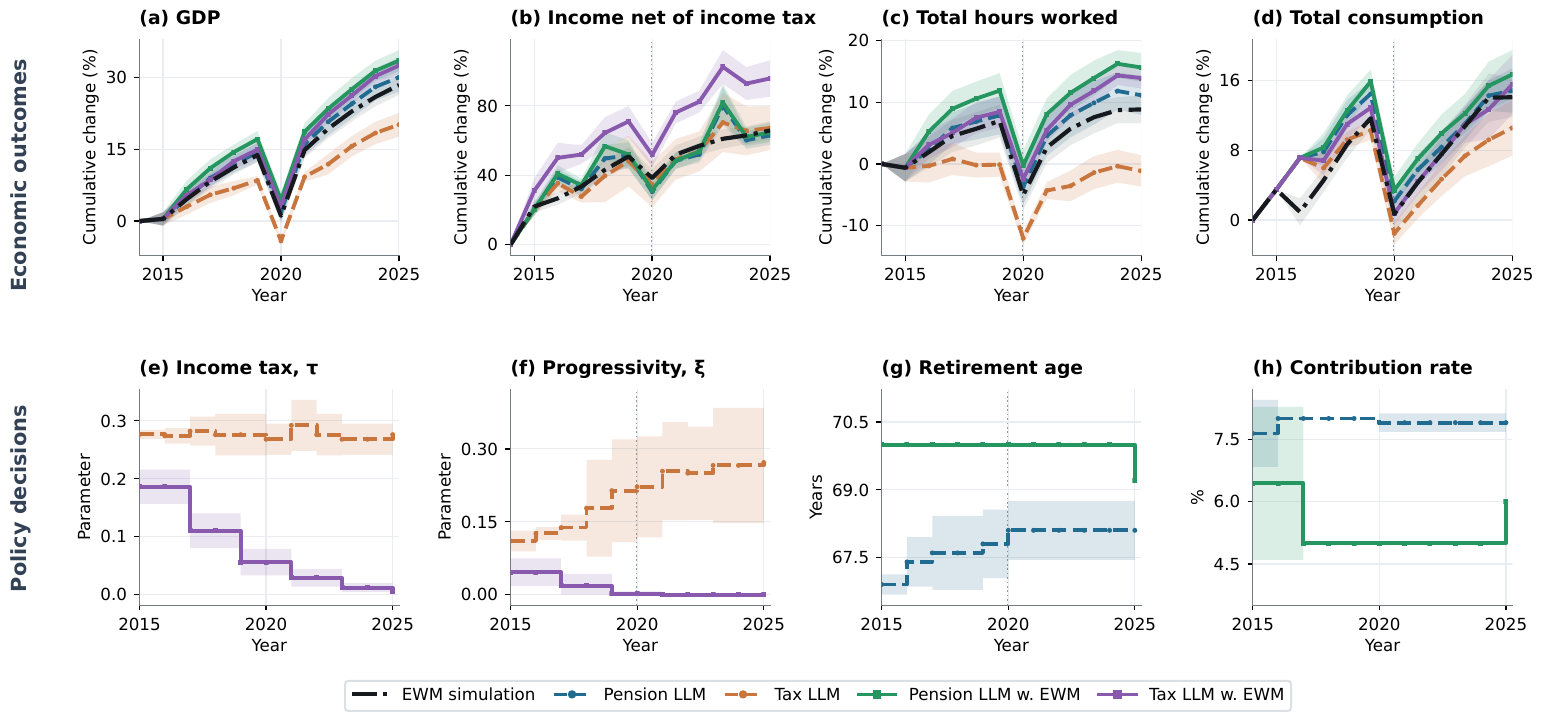}
\caption{Tax and pension policy tasks under the GDP objective with the 2020 epidemic shock. Panels and plotting conventions match Figure~\ref{fig:rq4-no-shock}; the vertical line marks 2020.}
\label{fig:rq4-shock}
\end{figure}

\textbf{LLM w. EWM achieved higher cumulative reward than Direct LLM in both policy tracks, with and without the epidemic shock} (Table~\ref{tab:rq4-performance}). Mean relative gains were $6.13\%$ without the shock and $6.55\%$ under the shock in the tax track, and $1.24\%$ and $1.71\%$ in the pension track, respectively. The advantage was larger on average under the shock in both tracks, indicating greater benefits from EWM-guided planning in disrupted conditions.

\textbf{LLM w. EWM also achieved higher simulated GDP, with the clearest transmission mechanism in the tax track} (Figures~\ref{fig:rq4-no-shock} and~\ref{fig:rq4-shock}). Tax LLM w. EWM generally selected lower income-tax and progressivity parameters, raising income net of income tax from 2015. Total hours worked rose relative to Direct LLM from 2016, consistent with stronger after-tax incentives to work. Within EWM, the increase in hours expanded labour input to firms and GDP, while consumption followed in 2017 as a delayed response to higher after-tax income. By 2025, total hours worked and GDP were $13.29\%$ and $8.89\%$ higher without the shock, and $15.28\%$ and $10.29\%$ higher under the shock, respectively.

\textbf{The pension track showed a similar but weaker labour-supply channel.} LLM w. EWM chose later retirement and, from 2017, generally lower contribution rates than Direct LLM, alongside more hours worked and higher firm labour input but only modest consumption gains. By 2025, hours were $3.10\%$ higher without the shock and $4.07\%$ higher under it, with GDP gains of $2.05\%$ and $2.69\%$, respectively. During the 2020 shock, LLM w. EWM reduced the tax-track GDP decline ($10.36\%$ versus $11.65\%$ for Direct LLM), while pension-track declines were similar ($10.89\%$ versus $10.82\%$; Figure~\ref{fig:rq4-shock}).

\FloatBarrier
\Needspace{.62\textheight}
\subsubsection{EWM-guided planning for welfare and pension sustainability}

\begin{figure}[H]
\centering
\includegraphics[width=\textwidth]{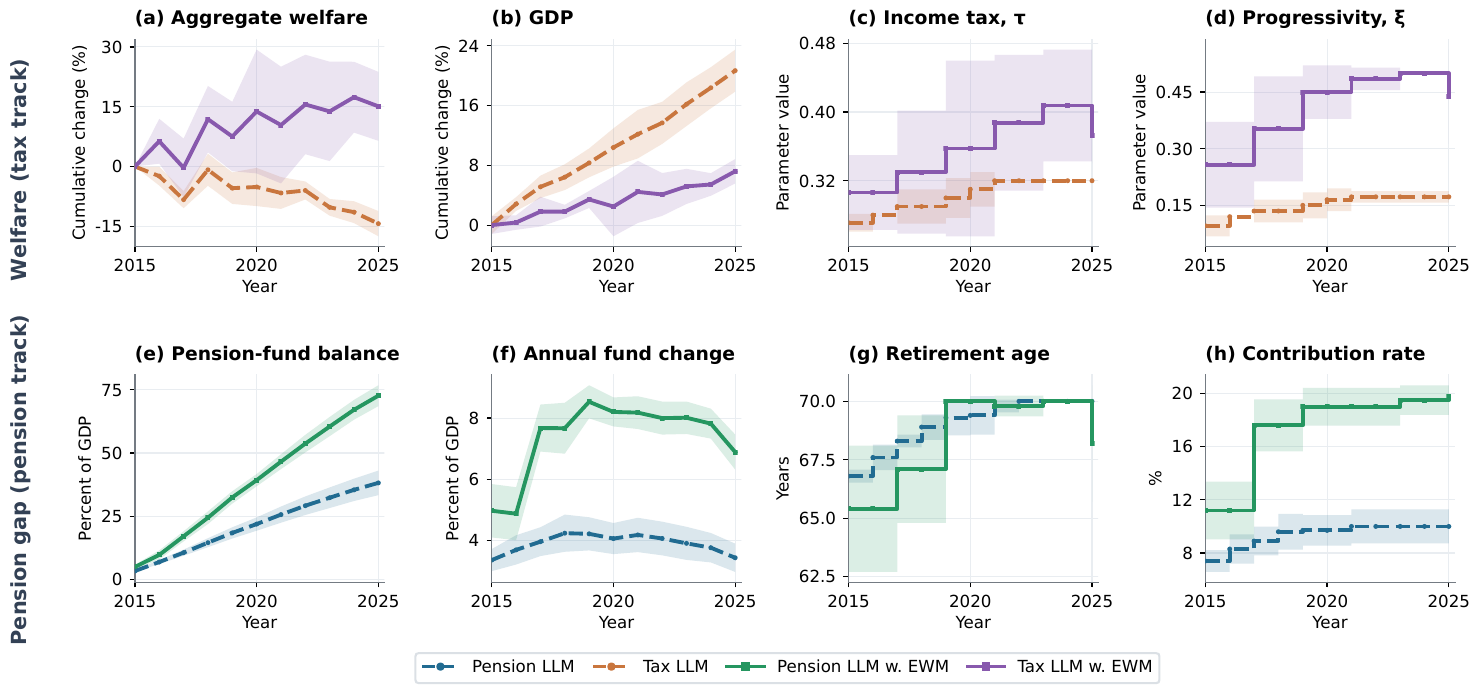}
\caption{Social-welfare objective via tax policy (upper row) and pension-fund accumulation objective via pension policy (lower row), without the epidemic shock. Welfare and GDP are shown as cumulative percentage changes from 2015 and 2014, respectively; the pension-fund balance and its annual change are expressed as percentages of GDP. Shading shows one sample standard deviation around the seed mean.}
\label{fig:rq4-objective-actions}
\end{figure}

\FloatBarrier
\Needspace{7\baselineskip}
\textbf{LLM w. EWM better optimized each specified objective, but exposed clear trade-offs with GDP} (Figure~\ref{fig:rq4-objective-actions}). Under the welfare objective, LLM w. EWM selected higher income-tax and progressivity parameters, increasing final-year aggregate welfare by $34.40\%$ while reducing total hours worked and GDP by $12.39\%$ and $11.11\%$, respectively. Under the pension-fund accumulation objective, it raised the mean contribution rate from $9.4\%$ to $17.5\%$, increasing the 2025 fund balance by $87.11\%$ while reducing hours and GDP by $3.41\%$ and $2.47\%$. Overall, evaluating candidate policies with EWM helped LLM agents make more economically informed decisions that better aligned with the specified policy objectives.

\section{Conclusion}

In this paper, we define an economic world model (EWM) as an agent's internal representation of how an economic system works. We then propose a general framework for building EWMs by decomposing the objective into three interconnected tasks: simulating agent behaviour, learning economic transitions and systematically evaluating the resulting dynamics. As an initial implementation, we train household policies on microeconomic data and calibrate a structured economic transition model on macroeconomic data, thereby linking individual decisions to aggregate dynamics. This initial EWM reproduces parts of the observed trajectories of economic variables and the short-run transmission from government borrowing to consumption in a benchmark life-cycle model. Together, these results show that the EWM framework can reproduce aspects of economic dynamics, while also revealing substantial room for improvement.

\Needspace{10\baselineskip}
We therefore hope to work with researchers across economics and AI to build a stronger EWM together. Economists can enrich the economic transition model with more accurate mechanisms and institutions or extend it to new settings. AI researchers can develop more powerful data-driven methods to better simulate agent behaviour and learn economic transitions. The micro-to-macro evaluation framework can assess these advances and provide feedback for further refinement. As these contributions accumulate, the shared EWM can become progressively more capable.

Looking ahead, we hope to build on mechanisms already established in economics and use AI's powerful computational capabilities to move one step further. This may enable EWMs not only to reproduce real economic dynamics and mechanisms more faithfully, but also to uncover new economic regularities that inform better decisions and benefit society.

\begingroup
\small
\setlength{\bibsep}{3pt plus 1pt minus 1pt}

\raggedright
\bibliography{references}

@article{haavelmo1944probability,
  author  = {Haavelmo, Trygve},
  title   = {The Probability Approach in Econometrics},
  journal = {Econometrica},
  volume  = {12},
  number  = {Supplement},
  pages   = {iii--vi, 1--115},
  year    = {1944},
  doi     = {10.2307/1906935}
}

@book{klein1955econometric,
  author = {Klein, Lawrence R. and Goldberger, Arthur S.},
  title = {An Econometric Model of the United States, 1929--1952},
  series = {Contributions to Economic Analysis},
  volume = {9},
  publisher = {North-Holland Publishing Company},
  address = {Amsterdam},
  year = {1955}
}

@incollection{bernanke1999financial,
  author = {Bernanke, Ben S. and Gertler, Mark and Gilchrist, Simon},
  title = {The Financial Accelerator in a Quantitative Business Cycle Framework},
  editor = {Taylor, John B. and Woodford, Michael},
  booktitle = {Handbook of Macroeconomics},
  volume = {1C},
  chapter = {21},
  pages = {1341--1393},
  publisher = {Elsevier},
  year = {1999},
  doi = {10.1016/S1574-0048(99)10034-X}
}

@article{huggett1993risk,
  title={The risk-free rate in heterogeneous-agent incomplete-insurance economies},
  author={Huggett, Mark},
  journal={Journal of economic Dynamics and Control},
  volume={17},
  number={5-6},
  pages={953--969},
  year={1993},
  publisher={Elsevier},
  doi = {10.1016/0165-1889(93)90024-M}
}

@article{krusell1998income,
  title={Income and wealth heterogeneity in the macroeconomy},
  author={Krusell, Per and Smith, Jr, Anthony A},
  journal={Journal of political Economy},
  volume={106},
  number={5},
  pages={867--896},
  year={1998},
  publisher={The University of Chicago Press}
}

@article{epstein1999agent,
  title={Agent-based computational models and generative social science},
  author={Epstein, Joshua M},
  journal={Complexity},
  volume={4},
  number={5},
  pages={41--60},
  year={1999},
  publisher={Wiley Online Library}
}

@article{dosi2010schumpeter,
  title={Schumpeter meeting Keynes: A policy-friendly model of endogenous growth and business cycles},
  author={Dosi, Giovanni and Fagiolo, Giorgio and Roventini, Andrea},
  journal={Journal of economic dynamics and control},
  volume={34},
  number={9},
  pages={1748--1767},
  year={2010},
  publisher={Elsevier}
}

@article{caiani2016agent,
  title={Agent based-stock flow consistent macroeconomics: Towards a benchmark model},
  author={Caiani, Alessandro and Godin, Antoine and Caverzasi, Eugenio and Gallegati, Mauro and Kinsella, Stephen and Stiglitz, Joseph E},
  journal={Journal of Economic Dynamics and Control},
  volume={69},
  pages={375--408},
  year={2016},
  publisher={Elsevier}
}

@article{stockwatson2002,
  author  = {James H. Stock and Mark W. Watson},
  title   = {Macroeconomic Forecasting Using Diffusion Indexes},
  journal = {Journal of Business \& Economic Statistics},
  volume  = {20},
  number  = {2},
  pages   = {147--162},
  year    = {2002},
  doi     = {10.1198/073500102317351921}
}

@article{sims1980,
  author  = {Christopher A. Sims},
  title   = {Macroeconomics and Reality},
  journal = {Econometrica},
  volume  = {48},
  number  = {1},
  pages   = {1--48},
  year    = {1980},
  doi     = {10.2307/1912017}
}

@incollection{delNegroSchorfheide2013,
  author    = {Del Negro, Marco and Schorfheide, Frank},
  title     = {{DSGE} Model-Based Forecasting},
  booktitle = {Handbook of Economic Forecasting},
  volume    = {2},
  pages     = {57--140},
  year      = {2013},
  publisher = {Elsevier},
  doi       = {10.1016/B978-0-444-53683-9.00002-5}
}

@misc{blsCEX2024,
  author       = {{U.S. Bureau of Labor Statistics}},
  title = {Consumer Expenditure Survey: Interview Survey Public Use Microdata, 2014--2024},
  howpublished = {U.S. Bureau of Labor Statistics},
  year         = {2024},
  url = {https://www.bls.gov/cex/pumd_data.htm},
  note = {Annual Interview Survey files for the stated survey years}
}

@misc{fredData2026,
  author       = {{Federal Reserve Bank of St. Louis}},
  title = {Macroeconomic Series from Federal Reserve Economic Data ({FRED})},
  howpublished = {Federal Reserve Bank of St. Louis},
  year         = {2026},
  url          = {https://fred.stlouisfed.org/},
  note = {{BEA}: GDPCA, PCECCA, GDP and GDPDEF. {BLS}: HOANBS and COMPNFB. {U.S. Treasury}: GFDEBTN. Revised data snapshot retrieved 12 October 2026}
}

@misc{fredDebtRatio2026,
  author       = {{Federal Reserve Bank of St. Louis}},
  title = {Federal Debt: Total Public Debt as Percent of Gross Domestic Product ({GFDEGDQ188S})},
  howpublished = {Archival Federal Reserve Economic Data ({ALFRED})},
  year         = {2026},
  url = {https://alfred.stlouisfed.org/graph/alfredgraph.csv?id=GFDEGDQ188S&cosd=2024-10-01&coed=2024-10-01&vintage_date=2026-11-10},
  note = {Vintage 10 November 2026. The 2024 Q4 observation is 121.01285 percent}
}

@misc{beaNationalAccounts2026,
  author       = {{U.S. Bureau of Economic Analysis}},
  title        = {National {GDP} and Personal Income Accounts},
  howpublished = {U.S. Bureau of Economic Analysis, Interactive Data Application},
  year         = {2026},
  url          = {https://www.bea.gov/itable/national-gdp-and-personal-income},
  note = {GDPCA, PCECCA, GDP and GDPDEF retrieved via FRED}
}

@misc{beaFixedAssets2026,
  author       = {{U.S. Bureau of Economic Analysis}},
  title        = {Fixed Assets Accounts},
  howpublished = {U.S. Bureau of Economic Analysis},
  year         = {2026},
  url          = {https://apps.bea.gov/national/FixedAssets/},
  note = {Section 1, private nonresidential fixed assets, rows of Tables 1.1 and 1.2 for current-cost net stock and the chain-type quantity index, respectively}
}

@misc{federalReserveDFA2026,
  author       = {{Board of Governors of the Federal Reserve System}},
  title        = {Distributional Financial Accounts},
  howpublished = {Financial Accounts of the United States},
  year         = {2026},
  url          = {https://www.federalreserve.gov/releases/z1/dataviz/dfa/},
  note = {Quarterly age-group household wealth accounts. The public age categories group ages 55--69 together}
}

@misc{ha2018world,
  author = {Ha, David and Schmidhuber, J{\"u}rgen},
  title = {World Models},
  year = {2018},
  eprint = {1803.10122},
  archivePrefix = {arXiv},
  primaryClass = {cs.LG},
  doi = {10.48550/arXiv.1803.10122},
  url = {https://arxiv.org/abs/1803.10122},
  howpublished = {{arXiv}:1803.10122}
}

@misc{lecun2022path,
  author = {LeCun, Yann},
  title = {A Path Towards Autonomous Machine Intelligence},
  year = {2022},
  howpublished = {OpenReview technical report},
  note = {Version 0.9.2, June 27, 2022},
  url = {https://openreview.net/pdf?id=BZ5a1r-kVsf}
}

@article{hafner2025mastering,
  author = {Hafner, Danijar and Pasukonis, Jurgis and Ba, Jimmy and Lillicrap, Timothy},
  title = {Mastering Diverse Control Tasks through World Models},
  journal = {Nature},
  volume = {640},
  number = {8059},
  pages = {647--653},
  year = {2025},
  doi = {10.1038/s41586-025-08744-2}
}

@misc{cong2025economic,
  author = {Cong, Lin William},
  title = {Economic World Models and Data-Driven Generative Equilibria},
  year = {2025},
  howpublished = {Cornell Legal Studies Research Paper (forthcoming)},
  doi = {10.2139/ssrn.6559940},
  url = {https://papers.ssrn.com/sol3/papers.cfm?abstract_id=6559940},
  note = {Written April 22, 2025; posted April 13, 2026; revised July 13, 2026}
}

@misc{han2026agenticeconomies,
  author = {Han, Jiale and Li, Xiang and Qian, Jing and Gu, Wenyuan and Gao, Pin and Luo, Ye and Zha, Hongyuan and Tao, Dacheng and Wang, Benyou and Cong, Lin William},
  title = {From Economic Agents to Agentic Economies: A Systems Blueprint for Economic World Models},
  year = {2026},
  eprint = {2608.06020v1},
  archivePrefix = {arXiv},
  primaryClass = {cs.AI},
  doi = {10.48550/arXiv.2608.06020},
  howpublished = {{arXiv}:2608.06020v1},
  url = {https://arxiv.org/abs/2608.06020v1}
}

@inproceedings{liu2022samplepomg,
  author = {Liu, Qinghua and Szepesv{\'a}ri, Csaba and Jin, Chi},
  title = {Sample-Efficient Reinforcement Learning of Partially Observable {Markov} Games},
  booktitle = {Advances in Neural Information Processing Systems},
  volume = {35},
  year = {2022},
  doi = {10.52202/068431-1330}
}

@inproceedings{mi2025econgym,
 author = {Mi, Qirui and Yang, Qipeng and Fan, Zijun and Fan, Wentian and Ma, Heyang and Ma, Chengdong and Xia, Siyu and An, Bo and Wang, Jun and Zhang, Haifeng},
 booktitle = {Advances in Neural Information Processing Systems},
 publisher = {Curran Associates, Inc.},
 title = {{EconGym}: A Scalable {AI} Testbed with Diverse Economic Tasks},
 volume = {38},
 year = {2025},
  note = {NeurIPS Datasets and Benchmarks Track},
  url = {https://proceedings.neurips.cc/paper_files/paper/2025/hash/40d45b1e23d00d5895e65778e85cf8ee-Abstract-Datasets_and_Benchmarks_Track.html}
}

@article{gertler1999government,
  author  = {Mark Gertler},
  title   = {Government Debt and Social Security in a Life-Cycle Economy},
  journal = {Carnegie-Rochester Conference Series on Public Policy},
  volume  = {50},
  pages   = {61--110},
  year    = {1999},
  doi = {10.1016/S0167-2231(99)00022-6}
}

@article{peyre2019optimal,
  author  = {Gabriel Peyr{\'e} and Marco Cuturi},
  title = {Computational Optimal Transport: With Applications to Data Science},
  journal = {Foundations and Trends in Machine Learning},
  volume  = {11},
  number  = {5--6},
  pages   = {355--607},
  year    = {2019},
  doi = {10.1561/2200000073}
}

@incollection{dawid2018agent,
  author = {Dawid, Herbert and Delli Gatti, Domenico},
  title = {Agent-Based Macroeconomics},
  editor = {Hommes, Cars H. and LeBaron, Blake},
  booktitle = {Handbook of Computational Economics},
  volume = {4},
  pages = {63--156},
  publisher = {Elsevier},
  year = {2018},
  doi = {10.1016/bs.hescom.2018.02.006}
}

@article{kaplan2018monetary,
  title = {Monetary Policy According to {HANK}},
  author={Kaplan, Greg and Moll, Benjamin and Violante, Giovanni L},
  journal={American Economic Review},
  volume={108},
  number={3},
  pages={697--743},
  year={2018},
  publisher = {American Economic Association},
  doi = {10.1257/aer.20160042}
}

@article{kydland1982time,
  author = {Kydland, Finn E. and Prescott, Edward C.},
  title = {Time to Build and Aggregate Fluctuations},
  journal = {Econometrica},
  volume = {50},
  number = {6},
  pages = {1345--1370},
  year = {1982},
  doi = {10.2307/1913386}
}

@inproceedings{Li2024EconAgent,
  author = {Li, Nian and Gao, Chen and Li, Mingyu and Li, Yong and Liao, Qingmin},
  title = {{EconAgent}: Large Language Model-Empowered Agents for Simulating Macroeconomic Activities},
  booktitle = {Proceedings of the 62nd Annual Meeting of the Association for Computational Linguistics (Volume 1: Long Papers)},
  pages = {15523--15536},
  year = {2024},
  publisher = {Association for Computational Linguistics},
  doi = {10.18653/v1/2024.acl-long.829},
  url = {https://aclanthology.org/2024.acl-long.829/}
}

@article{lucas1976econometric,
  author = {Lucas, Robert E., Jr.},
  title = {Econometric Policy Evaluation: A Critique},
  journal = {Carnegie-Rochester Conference Series on Public Policy},
  volume = {1},
  pages = {19--46},
  year = {1976},
  publisher = {Elsevier},
  doi = {10.1016/S0167-2231(76)80003-6}
}

@inproceedings{Mi2023TaxAI,
  author = {Mi, Qirui and Xia, Siyu and Song, Yan and Zhang, Haifeng and Zhu, Shenghao and Wang, Jun},
  title = {{TaxAI}: A Dynamic Economic Simulator and Benchmark for Multi-Agent Reinforcement Learning},
  booktitle = {Proceedings of the 23rd International Conference on Autonomous Agents and Multiagent Systems},
  pages = {1390--1399},
  year = {2024},
  publisher = {International Foundation for Autonomous Agents and Multiagent Systems},
  url = {https://www.ifaamas.org/Proceedings/aamas2024/pdfs/p1390.pdf}
}

@incollection{Mi2024StackelbergMFG,
  author = {Mi, Qirui and Zhao, Zhiyu and Ma, Chengdong and Xia, Siyu and Song, Yan and Yang, Mengyue and Wang, Jun and Zhang, Haifeng},
  title = {Learning Macroeconomic Policies Through Dynamic Stackelberg Mean-Field Games},
  booktitle = {ECAI 2025},
  series = {Frontiers in Artificial Intelligence and Applications},
  volume = {413},
  pages = {3655--3662},
  publisher = {IOS Press},
  year = {2025},
  doi = {10.3233/FAIA251243}
}

@article{schelling1971dynamic,
  title={Dynamic models of segregation},
  author={Schelling, Thomas C},
  journal={Journal of mathematical sociology},
  volume={1},
  number={2},
  pages={143--186},
  year={1971},
  publisher={Taylor \& Francis}
}

@article{smets2007shocks,
  title = {Shocks and Frictions in {US} Business Cycles: A Bayesian {DSGE} Approach},
  author={Smets, Frank and Wouters, Rafael},
  journal = {American Economic Review},
  volume={97},
  number={3},
  pages={586--606},
  year={2007},
  publisher={American Economic Association},
  doi = {10.1257/aer.97.3.586}
}

@book{Sutton2018,
  author = {Sutton, Richard S. and Barto, Andrew G.},
  title = {Reinforcement Learning: An Introduction},
  edition = {2},
  publisher = {MIT Press},
  address = {Cambridge, MA},
  year = {2018}
}

@book{tesfatsion2006handbook,
  title = {Handbook of Computational Economics: Agent-Based Computational Economics},
  editor={Tesfatsion, Leigh and Judd, Kenneth L},
  volume={2},
  year={2006},
  publisher={Elsevier},
  url = {https://shop.elsevier.com/books/handbook-of-computational-economics/tesfatsion/978-0-444-51253-6}
}

@article{tesfatsion2002bottomup,
  author  = {Tesfatsion, Leigh},
  title   = {Agent-Based Computational Economics: Growing Economies from the Bottom Up},
  journal = {Artificial Life},
  volume  = {8},
  number  = {1},
  pages   = {55--82},
  year    = {2002},
  doi     = {10.1162/106454602753694765}
}

@article{Zheng2022AIEconomist,
  author = {Zheng, Stephan and Trott, Alexander and Srinivasa, Sunil and Parkes, David C. and Socher, Richard},
  title = {The {AI} Economist: Taxation Policy Design via Two-Level Deep Multiagent Reinforcement Learning},
  journal = {Science Advances},
  volume = {8},
  number = {18},
  pages = {eabk2607},
  year = {2022},
  doi = {10.1126/sciadv.abk2607}
}

@article{aiyagari1994uninsured,
  title={Uninsured idiosyncratic risk and aggregate saving},
  author={Aiyagari, S Rao},
  journal={The Quarterly Journal of Economics},
  volume={109},
  number={3},
  pages={659--684},
  year={1994},
  publisher={MIT Press}
}

@techreport{Atashbar2023RLRBC,
  author = {Atashbar, Tohid and Shi, Rui Aruhan},
  title = {{AI} and Macroeconomic Modeling: Deep Reinforcement Learning in an {RBC} Model},
  institution = {International Monetary Fund},
  type = {Working Paper},
  number = {2023/040},
  year = {2023},
  doi = {10.5089/9798400235252.001},
  url = {https://doi.org/10.5089/9798400235252.001}
}

@misc{Douglas2024ConsumptionSavings,
  author = {Douglas, Michael R. and Verstyuk, Sergiy},
  title = {Consumption and Savings with Large Language Model Agents},
  year = {2024},
  doi = {10.2139/ssrn.4909749},
  howpublished = {SSRN working paper},
  url = {https://papers.ssrn.com/sol3/papers.cfm?abstract_id=4909749},
  note = {2024 version of SSRN record 4909749}
}

@misc{Fish2024AlgorithmicCollusion,
  author = {Fish, Sara and Gonczarowski, Yannai A. and Shorrer, Ran I.},
  title = {Algorithmic Collusion by Large Language Models},
  year = {2026},
  eprint = {2404.00806v5},
  archivePrefix = {arXiv},
  primaryClass = {econ.GN},
  note = {Preprint revised 5 March 2026},
  howpublished = {{arXiv}:2404.00806v5},
  url = {https://arxiv.org/abs/2404.00806v5}
}

@misc{Gao2024SimulatingFinancialMarket,
  author = {Gao, Shen and Wen, Yuntao and Zhu, Minghang and Wei, Jianing and Cheng, Yuhan and Zhang, Qunzi and Shang, Shuo},
  title = {Simulating Financial Market via Large Language Model Based Agents},
  year = {2024},
  eprint = {2406.19966},
  archivePrefix = {arXiv},
  primaryClass = {cs.CL},
  howpublished = {{arXiv}:2406.19966},
  url = {https://arxiv.org/abs/2406.19966}
}

@misc{Han2024GuineaPig,
  author = {Han, Xu and Wu, Zengqing and Xiao, Chuan},
  title = {``Guinea Pig Trials'' Utilizing {GPT}: A Novel Smart Agent-Based Modeling Approach for Studying Firm Competition and Collusion},
  year = {2024},
  eprint = {2308.10974v4},
  archivePrefix = {arXiv},
  primaryClass = {cs.AI},
  note = {Preprint revised 31 January 2024},
  howpublished = {{arXiv}:2308.10974v4},
  url = {https://arxiv.org/abs/2308.10974v4}
}

@misc{Hill2021SolvingHeterogeneousGE,
  author = {Hill, Edward and Bardoscia, Marco and Turrell, Arthur},
  title = {Solving Heterogeneous General Equilibrium Economic Models with Deep Reinforcement Learning},
  year = {2021},
  eprint = {2103.16977},
  archivePrefix = {arXiv},
  primaryClass = {econ.GN},
  howpublished = {{arXiv}:2103.16977},
  url = {https://arxiv.org/abs/2103.16977}
}

@misc{Karten2025LLMEconomist,
  author = {Karten, Seth and Li, Wenzhe and Ding, Zihan and Kleiner, Samuel and Bai, Yu and Jin, Chi},
  title = {{LLM} Economist: Large Population Models and Mechanism Design in Multi-Agent Generative Simulacra},
  year = {2025},
  eprint = {2507.15815},
  archivePrefix = {arXiv},
  primaryClass = {cs.MA},
  howpublished = {{arXiv}:2507.15815},
  url = {https://arxiv.org/abs/2507.15815}
}

@article{Koster2022DemocraticAI,
  author = {Koster, Raphael and Balaguer, Jan and Tacchetti, Andrea and Weinstein, Ari and Zhu, Tina and Hauser, Oliver and Williams, Duncan and Campbell-Gillingham, Lucy and Thacker, Phoebe and Botvinick, Matthew and Summerfield, Christopher},
  title = {Human-Centred Mechanism Design with {Democratic AI}},
  journal = {Nature Human Behaviour},
  volume = {6},
  number = {10},
  pages = {1398--1407},
  year = {2022},
  doi = {10.1038/s41562-022-01383-x}
}

@inproceedings{Park2023GenerativeAgents,
  author = {Park, Joon Sung and O'Brien, Joseph C. and Cai, Carrie J. and Morris, Meredith Ringel and Liang, Percy and Bernstein, Michael S.},
  title = {Generative Agents: Interactive Simulacra of Human Behavior},
  booktitle = {Proceedings of the 36th Annual ACM Symposium on User Interface Software and Technology},
  pages = {1--22},
  year = {2023},
  publisher = {ACM},
  doi = {10.1145/3586183.3606763}
}

@misc{Shi2022LearningFromZero,
  author = {Shi, Rui Aruhan},
  title = {Learning from Zero: How to Make Consumption-Saving Decisions in a Stochastic Environment with an AI Algorithm},
  year = {2022},
  eprint = {2105.10099v2},
  archivePrefix = {arXiv},
  primaryClass = {econ.TH},
  note = {Preprint revised 18 February 2022},
  howpublished = {{arXiv}:2105.10099v2},
  url = {https://arxiv.org/abs/2105.10099v2}
}

@misc{Trott2021BuildingFoundation,
  author = {Trott, Alexander and Srinivasa, Sunil and van der Wal, Douwe and Haneuse, Sebastien and Zheng, Stephan},
  title = {Building a Foundation for Data-Driven, Interpretable, and Robust Policy Design Using the {AI} Economist},
  year = {2021},
  eprint = {2108.02904},
  archivePrefix = {arXiv},
  primaryClass = {cs.LG},
  howpublished = {{arXiv}:2108.02904},
  url = {https://arxiv.org/abs/2108.02904}
}

@inproceedings{Xu2023RegularizedCompetitiveEquilibria,
  author = {Xu, Ruitu and Min, Yifei and Wang, Tianhao and Jordan, Michael I. and Wang, Zhaoran and Yang, Zhuoran},
  title = {Finding Regularized Competitive Equilibria of Heterogeneous Agent Macroeconomic Models via Reinforcement Learning},
  booktitle = {Proceedings of The 26th International Conference on Artificial Intelligence and Statistics},
  series = {Proceedings of Machine Learning Research},
  volume = {206},
  pages = {375--407},
  year = {2023},
  publisher = {PMLR}
}

@article{calvano2020artificial,
Author = {Calvano, Emilio and Calzolari, Giacomo and Denicolò, Vincenzo and Pastorello, Sergio},
Title = {Artificial Intelligence, Algorithmic Pricing, and Collusion},
Journal = {American Economic Review},
Volume = {110},
Number = {10},
Year = {2020},
Month = {October},
Pages = {3267--3297},
DOI = {10.1257/aer.20190623}}

@article{stockwatson1996structural,
  author  = {Stock, James H. and Watson, Mark W.},
  title   = {Evidence on Structural Instability in Macroeconomic Time Series Relations},
  journal = {Journal of Business \& Economic Statistics},
  volume  = {14},
  number  = {1},
  pages   = {11--30},
  year    = {1996},
  doi     = {10.1080/07350015.1996.10524626}
}

@article{burr2024transmission,
  author = {Burr, Natalie and Willems, Tim},
  title = {About a Rate of (General) Interest: How Monetary Policy Transmits},
  journal = {Bank of England Quarterly Bulletin},
  year = {2024},
  month = jul,
  url = {https://www.bankofengland.co.uk/quarterly-bulletin/2024/2024/about-a-rate-of-general-interest-how-monetary-policy-transmits}
}

@misc{federalReserveGS1,
  author       = {{Board of Governors of the Federal Reserve System}},
  year         = {2026},
  title = {Market Yield on {U.S.} Treasury Securities at 1-Year Constant Maturity, Quoted on an Investment Basis ({GS1})},
  howpublished = {H.15 Selected Interest Rates, retrieved from FRED},
  url          = {https://fred.stlouisfed.org/series/GS1},
  note = {Monthly nominal yield. The fixed-origin experiment uses annual means in the calibration window and the December 2019 rate in the test years}
}

@article{fujiwara2008lifecycle,
  author  = {Ippei Fujiwara and Yuki Teranishi},
  title = {A Dynamic New {Keynesian} Life-Cycle Model: Societal Aging, Demographics, and Monetary Policy},
  journal = {Journal of Economic Dynamics and Control},
  volume  = {32},
  number  = {8},
  pages   = {2398--2427},
  year    = {2008},
  doi     = {10.1016/j.jedc.2007.09.002}
}

@inproceedings{ross2011reduction,
  author = {Ross, Stephane and Gordon, Geoffrey and Bagnell, Drew},
  title = {A Reduction of Imitation Learning and Structured Prediction to No-Regret Online Learning},
  booktitle = {Proceedings of the Fourteenth International Conference on Artificial Intelligence and Statistics},
  series = {Proceedings of Machine Learning Research},
  volume = {15},
  pages = {627--635},
  publisher = {PMLR},
  year = {2011},
  url = {https://proceedings.mlr.press/v15/ross11a.html}
}

@article{sobol1967distribution,
  author = {Sobol', I. M.},
  title = {On the Distribution of Points in a Cube and the Approximate Evaluation of Integrals},
  journal = {USSR Computational Mathematics and Mathematical Physics},
  volume = {7},
  number = {4},
  pages = {86--112},
  year = {1967},
  doi = {10.1016/0041-5553(67)90144-9}
}

@article{regis2007stochastic,
  author = {Regis, Rommel G. and Shoemaker, Christine A.},
  title = {A Stochastic Radial Basis Function Method for the Global Optimization of Expensive Functions},
  journal = {INFORMS Journal on Computing},
  volume = {19},
  number = {4},
  pages = {497--509},
  year = {2007},
  doi = {10.1287/ijoc.1060.0182}
}
\endgroup
\end{document}